\def\year{2021}\relax
\documentclass[letterpaper]{article} 
\usepackage{aaai21}  
\usepackage{times}  
\usepackage{helvet} 
\usepackage{courier}  
\usepackage[hyphens]{url}  
\usepackage{graphicx} 
\usepackage{graphicx}  
\usepackage{natbib}  
\usepackage{caption} 

\usepackage{amsmath}
\usepackage{multirow}
\usepackage{makecell}
\usepackage{amssymb}
\usepackage{dsfont}
\usepackage{amssymb}
\usepackage{color, colortbl}
\usepackage{pifont}
\usepackage{hhline}
\usepackage{subfigure}
\usepackage{booktabs}

\title{ePointDA: An End-to-End Simulation-to-Real Domain Adaptation Framework for LiDAR Point Cloud Segmentation}
\author{Sicheng Zhao$^{1}$\thanks{Corresponding Author. $^{\#}$Equal contribution.}$^{\#}$, Yezhen Wang$^{23\#}$, Bo Li$^{1}$, Bichen Wu$^{4}$, Yang Gao$^{5}$, Pengfei Xu$^2$,\\
Trevor Darrell$^{1}$, Kurt Keutzer$^{1}$\\}
\affiliations{
$^1$University of California, Berkeley\ $^2$Didi Chuxing\ 
$^3$ University of California, San Diego\ 
$^4$ Facebook Inc\ $^5$Tsinghua University\\
\{schzhao,yezhen.wang0305,drluodian\}@gmail.com, wbc@fb.com, gaoyangiiis@tsinghua.edu.cn, xupengfeipf@didichuxing.com, trevor@eecs.berkeley.edu, keutzer@berkeley.edu
}

\begin{document}

\maketitle

\begin{abstract}
Due to its robust and precise distance measurements, LiDAR plays an important role in scene understanding for autonomous driving. Training deep neural networks (DNNs) on LiDAR data requires large-scale point-wise annotations, which are time-consuming and expensive to obtain. Instead, simulation-to-real domain adaptation (SRDA) trains a DNN using unlimited synthetic data with automatically generated labels and transfers the learned model to real scenarios. Existing SRDA methods for LiDAR point cloud segmentation mainly employ a multi-stage pipeline and focus on feature-level alignment. They require prior knowledge of real-world statistics and ignore the pixel-level dropout noise gap and the spatial feature gap between different domains. In this paper, we propose a novel end-to-end framework, named ePointDA, to address the above issues. Specifically, ePointDA consists of three modules: self-supervised dropout noise rendering, statistics-invariant and spatially-adaptive feature alignment, and transferable segmentation learning. The joint optimization enables ePointDA to bridge the domain shift at the pixel-level by explicitly rendering dropout noise for synthetic LiDAR and at the feature-level by spatially aligning the features between different domains, without requiring the real-world statistics. Extensive experiments adapting from synthetic GTA-LiDAR to real KITTI and SemanticKITTI demonstrate the superiority of ePointDA for LiDAR point cloud segmentation.
\end{abstract}

\section{Introduction}
Many types of multimedia data, such as images captured by cameras and point clouds collected by LiDAR (Light Detection And Ranging) and RADAR (RAdio Detection And Ranging) can help to understand the semantics of complex scenes for autonomous driving.
Among these sensors, LiDAR is an essential one for its specific properties~\cite{wu2018squeezeseg}. LiDAR can provide precise distance measurements; for example, the error of Velodyne HDL-64E is less than 2cm\footnote{\url{https://velodyneLiDAR.com/products/hdl-64e}}. Further, it is more robust to ambient lighting conditions (\textit{e.g.} night and shadow) than cameras and can obtain higher resolution and field of view than RADAR.

\begin{figure*}
\centering
\includegraphics[width=1.0\linewidth]{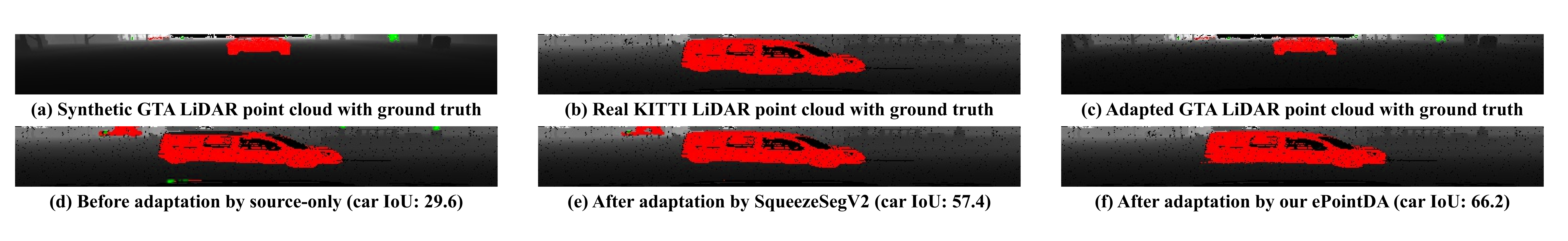}
\caption{An example of \emph{domain shift} between synthetic and real LiDAR point clouds, which are projected to a spherical surface for visualization (car in red, pedestrian in green). We can clearly see that: First, real point clouds (b) contain a lot of dropout noise (missing points) while the synthetic ones (a) do not; Second, the proposed ePointDA framework (f) significantly improves the domain adaptation performance for point-wise segmentation as compared to source-only (d) and SqueezeSegV2~\cite{wu2019squeezesegv2} (e); Finally, compared to (a), the adapted point clouds (c) with rendered dropout noise look more similar to the real ones.}
\label{fig:DomainShift}
\end{figure*}

Recent research has shown that deep neural networks (DNNs) can achieve state-of-the-art performance for point cloud classification and segmentation~\cite{qi2017pointnet,qi2017pointnet++,wu2018squeezeseg,wu2019squeezesegv2,zhang2019shellnet} with large-scale labeled data, which is usually time-consuming and expensive to obtain~\cite{wang2019latte}. However, unlimited \textit{synthetic} labeled data can be created using advanced simulators, such as CARLA\footnote{\url{http://www.carla.org}} and GTA-V\footnote{\url{https://www.rockstargames.com/V}} for autonomous driving. Unfortunately, due to the presence of domain shift between simulation and the real-world~\cite{wu2019squeezesegv2}, as shown in Figure~\ref{fig:DomainShift}, direct transfer often results in significant performance decay. Domain adaptation (DA) aims to learn a transferable model to minimize the impact of domain shift between the source and target domains~\cite{patel2015visual,zhao2020review}.



As the only simulation-to-real domain adaptation (SRDA) method for LiDAR point cloud segmentation, SqueezeSegV2~\cite{wu2019squeezesegv2} consists of three stages: learned intensity rendering, Geodesic correlation alignment, and progressive domain calibration. Although it achieved state-of-the-art SRDA performance at the time, there are some limitations. First, it employs a multi-stage pipeline and cannot be trained end-to-end. Second, it does not consider the pixel-level dropout noise gap between different domains. Third, the progressive calibration is inefficient and lacks of robustness, as the accurate real-world statistics is difficult to estimate and is evolving with incremental data. Fourth, the standard convolution in the segmentation model neglects the drastic difference between spatial features and corresponding spatial feature gap across domains.

One might argue that we can apply the DA methods for RGB image segmentation, especially the ones performing both feature-level and pixel-level alignments (\textit{e.g.} GTA-GAN~\cite{sankaranarayanan2018generate}, CyCADA~\cite{hoffman2018CyCADA}), to the SRDA problem for LiDAR point cloud segmentation. However, the 2D LiDAR images generated from 3D LiDAR point clouds projected onto a spherical surface~\cite{wu2018squeezeseg,wu2019squeezesegv2} are significantly different from RGB images. For example, RGB images  mainly consist of color and texture, the style of which can be well translated by Generative Adversarial Network (GAN)~\cite{goodfellow2014generative} and CycleGAN~\cite{zhu2017unpaired}; 2D LiDAR images are mainly about geometric information with dropout noise as the major domain gap between the synthetic and real data, as shown in Figure~\ref{fig:DomainShift}. Therefore, existing GAN-based DA methods usually do not perform well for LiDAR point cloud segmentation (see experiment for details).

In this paper, we design a novel end-to-end framework, ePointDA, to address the above issues in SRDA for LiDAR point cloud segmentation. First, we render the dropout noise for synthetic data based on a self-supervised model trained  on unlabeled real data, taking point coordinates as input and dropout noise as predictions. Second, we align the features of the simulation and real domains based on higher-order moment matching~\cite{chen2020homm} with statistics-invariant normalized features by instance normalization~\cite{ulyanov2016instance} and domain-invariant spatial attention by improving spatially-adaptive convolution~\cite{xu2020squeezesegv3}. The specific feature alignment method not only helps bridge the spatial feature gap, but also does not require prior access to sufficient real data to obtain the statistics, allowing it to deal better with the incremental real data and thus making it more robust and practical. Finally, we learn a transferable segmentation model based on the adapted images and corresponding synthetic labels.

In summary, the contributions of this paper are threefold:
\begin{itemize}
\item We are the first to study the simulation-to-real domain adaptation (SRDA) problem for LiDAR point cloud segmentation in an end-to-end manner.
\item We design a novel framework, named ePointDA, to bridge the domain gap between the simulation and real domains at both the pixel-level and the feature-level through self-supervised dropout noise rendering and statistics-invariant and spatially-adaptive feature alignment.
\item We conduct extensive experiments from synthetic GTA-LiDAR~\cite{wu2019squeezesegv2} to real KITTI~\cite{geiger2012we} and SemanticKITTI~\cite{behley2019semantickitti}, and respectively achieve 8.8\% and 7.5\% better IoU scores (on the ``car'' class) than the best DA baseline.
\end{itemize}

\section{Related Work}

\noindent\textbf{Point Cloud Segmentation.} Recent efforts on point cloud segmentation are typically based on DNNs. One straightforward way is to use the raw, un-ordered point clouds as input to a DNN. To deal with the order missing problem, symmetrical operators are usually applied, such as in PointNet~\cite{qi2017pointnet}, PointNet++~\cite{qi2017pointnet++}, and their improvements on hierarchical architecture~\cite{klokov2017escape}, sampling~\cite{dovrat2019learning}, reordering~\cite{li2018pointcnn}, grouping~\cite{li2018so}, and efficiency~\cite{liu2019point,liu2019densepoint,zhang2019shellnet}. There are also methods converting point clouds to regular 3D voxel grids~\cite{wang2017cnn,huang2018recurrent,le2018pointgrid,lei2019octree,mao2019interpolated,meng2019vv} or constructing graphs from point clouds for network processing~\cite{te2018rgcnn,jiang2019hierarchical,xu2018spidercnn,landrieu2018large,wang2019graph,wang2019dynamic}. However, these methods suffer from some limitations, such as inefficiency and point collision~\cite{lyu2020learning}. To address the efficiency problem and enable real-time inference, one popular method is to project 3D point clouds to 2D images, including sphere mapping~\cite{wu2018squeezeseg,wu2019squeezesegv2,milioto2019rangenet++,behley2019semantickitti,xu2020squeezesegv3}, 2D grid sampling~\cite{caltagirone2017fast}, and graph drawing~\cite{lyu2020learning}. In this paper, we follow the spherical projection method of SqueezeSeg~\cite{wu2018squeezeseg,wu2019squeezesegv2}.

\begin{figure*}
\centering
\includegraphics[width=1.0\linewidth]{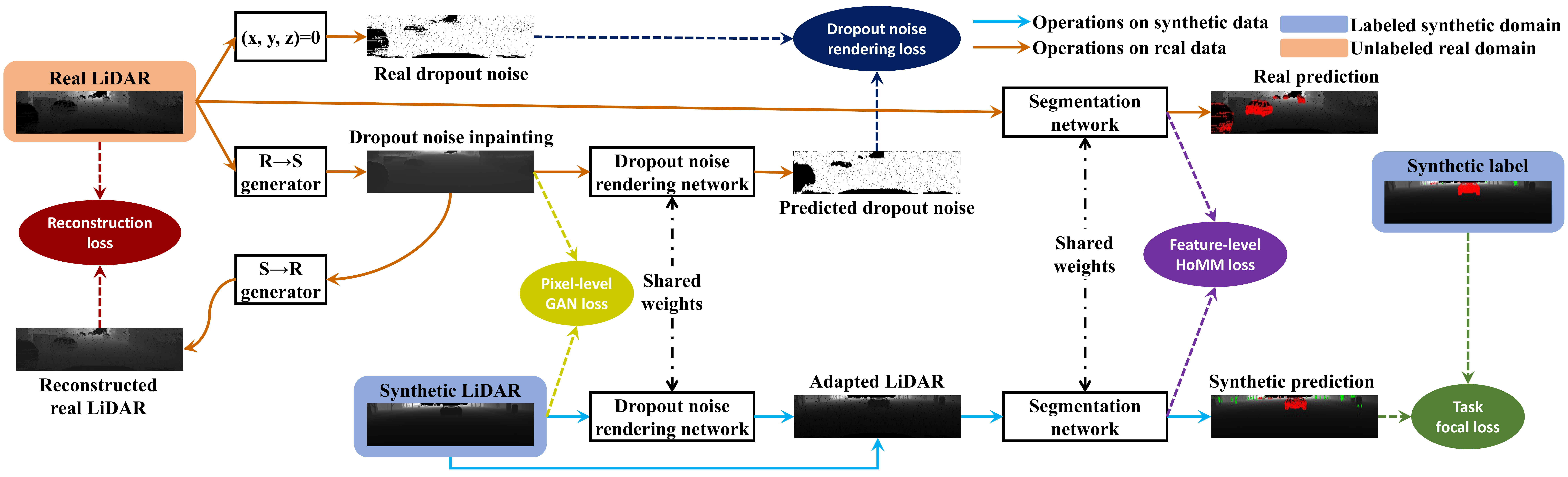}
\caption{The proposed SRDA framework ePointDA for LiDAR point cloud segmentation. The colored dashed arrows correspond to different losses. For clarity the real-to-simulation cycle is omitted. See Figure~\ref{fig:segmentationNetwork} for detailed segmentation network.}
\label{fig:Framework}
\end{figure*}

\noindent\textbf{Point Cloud Simulation.} Some efforts have been dedicated to creating large-scale real-world point cloud datasets, such as 3D bounding box to point-wise labeling~\cite{wang2019latte} and densely annotated SemanticKITTI dataset~\cite{behley2019semantickitti}. However, it is still difficult or impossible to collect all required point cloud scenes, such as traffic accidents in autonomous driving. The synthetic data generated by advanced simulators can achieve this goal with unlimited labeled data~\cite{richter2016playing,dosovitskiy2017carla,yue2018LiDAR,krahenbuhl2018free,tripathi2019learning}. In this paper, we employ the synthetic GTA-LiDAR dataset~\cite{wu2019squeezesegv2} with depth segmentation map generated by~\cite{krahenbuhl2018free} and Image-LiDAR registration in GTA-V by~\cite{yue2018LiDAR}.

\noindent\textbf{Unsupervised Domain Adaptation.} Most existing research on DA focuses on the single-source and unsupervised setting, \textit{i.e.} adapting from one labeled source domain to another unlabeled target domain~\cite{zhao2020review}. Recent deep unsupervised DA methods usually employ a conjoined architecture with two streams. Besides the task loss on the labeled source domain, another alignment loss is designed to align the source and target domains, such as discrepancy loss~\cite{long2015learning,sun2016return,zhuo2017deep,wu2019squeezesegv2,chen2020homm}, adversarial loss~\cite{tzeng2017adversarial,shrivastava2017learning,russo2018source,sankaranarayanan2018generate,zhao2018emotiongan,hoffman2018CyCADA,zhao2019cycleemotiongan,lee2019drop}, and self-supervision loss~\cite{sun2019unsupervised,carlucci2019domain,feng2019self,achituve2021self}. There are also some multi-source DA methods based on deep architectures~\cite{zhao2018adversarial,peng2019moment,zhao2019multi,zhao2020multi,lin2020multi,zhao2021curriculum}.

\begin{figure}
\centering
\includegraphics[width=1.0\linewidth]{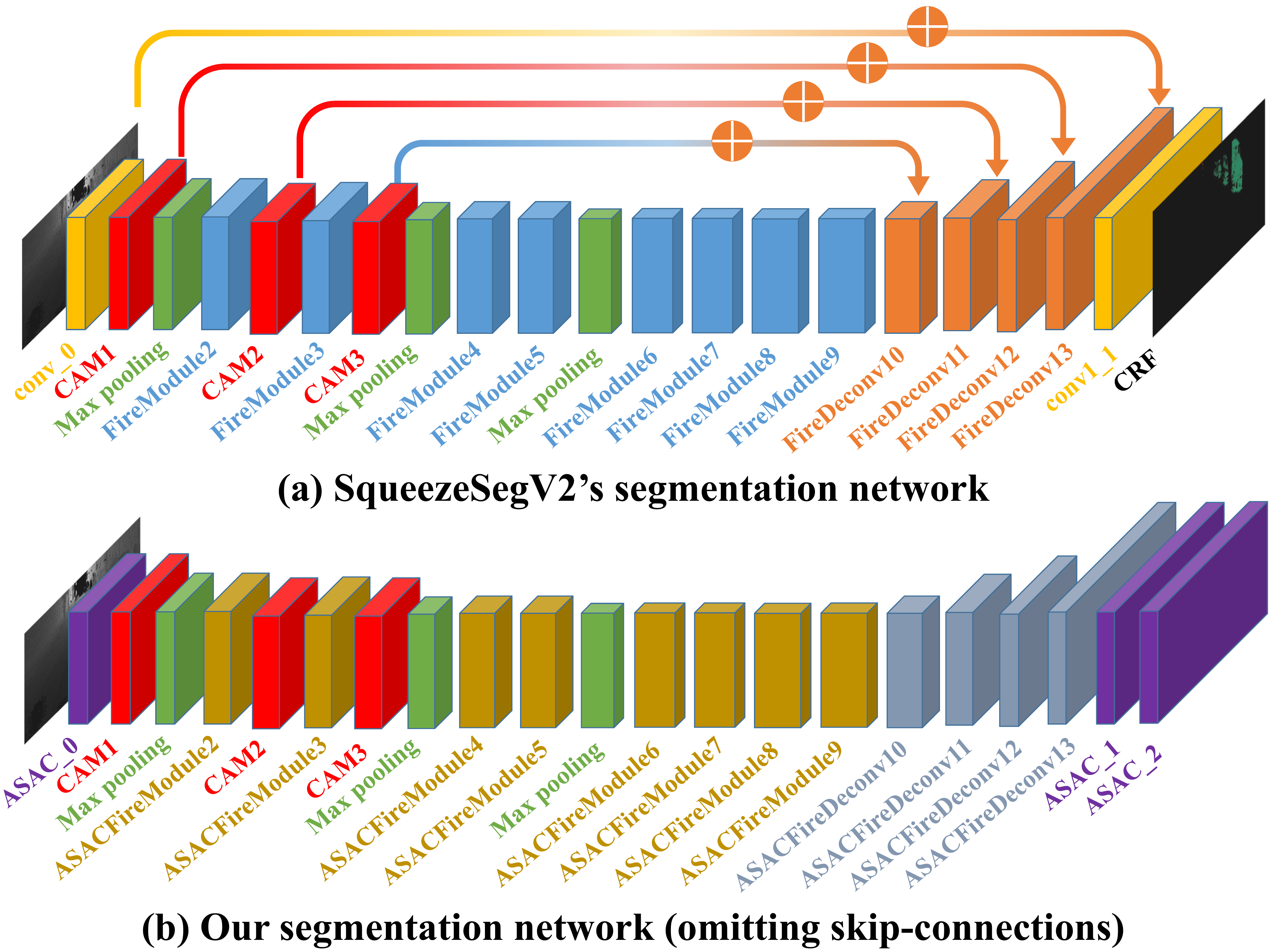}
\caption{Our segmentation network vs.  SqueezeSegV2~\cite{wu2019squeezesegv2}. We replace all standard convolution (conv) and the final conditional random field (CRF) with aligned spatially-adaptive convolution (ASAC). We replace all batch normalization after each conv with instance normalization.}
\label{fig:segmentationNetwork}
\end{figure}

On one hand, most SRDA methods exploring synthetic data focus on 2D RGB images for object classification~\cite{peng2019moment}, pose estimation~\cite{shrivastava2017learning}, and semantic segmentation~\cite{sankaranarayanan2018generate,zhao2019multi}. On the other hand, existing DA methods for LiDAR point cloud perception either conduct classification~\cite{qin2019pointdan,achituve2021self} or detection~\cite{saleh2019domain,rist2019cross} tasks, or perform real-to-real segmentation~\cite{rist2019cross,jiang2020LiDARnet}. The only SRDA method for LiDAR point cloud segmentation is SqueezeSegV2~\cite{wu2019squeezesegv2}, but it is trained stage by stage. We propose to study SRDA for LiDAR point cloud segmentation in an end-to-end manner.

\section{Approach}
\label{sec:Approach}
Given labeled synthetic LiDAR and unlabeled real LiDAR, our goal is to learn a transferable segmentation model by aligning the source simulation domain and target real domain. Following SqueezeSeg~\cite{wu2018squeezeseg,wu2019squeezesegv2}, we project sparse 3D LiDAR point clouds to 2D images for efficient processing, \textit{i.e.} projecting each point in the Cartesian coordinate to the angular coordinate. In this way, a LiDAR point cloud is transformed to a LiDAR image with size $H\times W\times C$, where $H$, $W$ are the height and width of the projected image\footnote{We use the LiDAR collected by Velodyne HDL-64E with 64 vertical channels, $H=64$; and use the frontal
90 degrees of the scan, dividing it into 512 grids, $W=512$.}, and $C$ is the number of image channels\footnote{In experiment, we use the Cartesian coordinates $(x,y,z)$ as features for each point, \textit{i.e.} $C=3$. We also tried other features, such as range and (rendered) intensity~\cite{wu2019squeezesegv2}, but the experiments show that adding these channels does not result in performance improvement for domain adaptation.}.

We consider the one-source, unsupervised, homogeneous, and closed-set SRDA scenario for LiDAR point cloud segmentation. That is, there is one labeled simulation domain and one unlabeled real domain, the observed LiDAR data of different domains are from the same space, and the label categories are shared across different domains. Suppose the projected synthetic images and corresponding labels drawn from the synthetic distribution $P_s(\mathbf{x},\mathbf{y})$ are $\mathbf{X}_s=\{\mathbf{x}_s^i\}_{i=1}^{N_s}$ and $\mathbf{Y}_s=\{\mathbf{y}_s^i\}_{i=1}^{N_s}$, respectively, where $\mathbf{x}_s^i\in \mathbb{R}^{H\times W\times C}$, $\mathbf{y}_s^i\in \{1,2,\cdots,L\}^{H\times W}$, $L$ is the number of label categories, and $N_s$ is the number of synthetic samples. Let $\mathbf{X}_r=\{\mathbf{x}_r^j\}_{j=1}^{N_r}$ denote the projected real images drawn from the real distribution $P_r(\mathbf{x})$, where $N_r$ is the number of real samples. On the basis of covariate shift and concept drift~\cite{patel2015visual}, we aim to learn a segmentation model that can correctly predict the labels for each pixel of a real sample trained on $\{(\mathbf{X}_s,\mathbf{Y}_s)\}$ and $\{\mathbf{X}_r\}$.

The framework of ePointDA is illustrated in Figure~\ref{fig:Framework}, which includes three modules. Self-supervised dropout noise rendering (SDNR) aims to bridge the domain shift at the pixel-level by generating adapted images based on the rendered dropout noise. Statistics-invariant and spatially-adaptive feature alignment aims to bridge the domain shift at the feature-level by considering the instance-wise statistics variations and spatial statistics differences. Transferable segmentation learning can then learn a transferable segmentation model based on the adapted images and corresponding synthetic labels.

\subsection{Self-Supervised Dropout Noise Rendering}
\label{ssec:DropoutNoiseRendering}

LiDAR point clouds in the real-world usually contain significant dropout noise, \textit{i.e.} missing points, where all coordinates $(x,y,z)$ are zero. However, synthetic LiDAR does not contain such noise, as it is difficult to simulate. Besides the random dropout noise, we propose an inpainting-based rendering method in a self-supervised manner to render other dropout noises, such as the ones caused by mirror reflection. 



First, we employ CycleGAN~\cite{zhu2017unpaired} to fill the dropout noise with the following pixel-level GAN loss and cycle-consistency loss:
\begin{equation}
\small
\begin{aligned}
\mathcal{L}_{GAN}^{r\rightarrow s}(G_{s},D_s, \mathbf{X}_r, \mathbf{X}_s)&=\mathbb{E}_{\mathbf{x}_{r}\sim \mathbf{X}_r}\log D_s(G_{s}(\mathbf{x}_{r}))\\
&+\mathbb{E}_{\mathbf{x}_s\sim \mathbf{X}_s}\log [1-D_s(\mathbf{x}_s)],
\end{aligned}
\label{equ:ganRS}
\end{equation}
\begin{equation}
\small
\begin{aligned}
\mathcal{L}_{GAN}^{s\rightarrow r}(G_{r},D_r, \mathbf{X}_s, \mathbf{X}_r)&=\mathbb{E}_{\mathbf{x}_{s}\sim \mathbf{X}_s}\log D_r(G_{r}(\mathbf{x}_{s}))\\
&+\mathbb{E}_{\mathbf{x}_r\sim \mathbf{X}_r}\log [1-D_r(\mathbf{x}_r)],
\end{aligned}
\label{equ:ganSR}
\end{equation}
\begin{equation}
\small
\begin{aligned}
\mathcal{L}_{cyc}(G_{s},G_{r},\mathbf{X}_{r}, \mathbf{X}_s)&=\mathbb{E}_{\mathbf{x}_{r}\sim \mathbf{X}_{r}}\parallel G_{r}(G_{s}(\mathbf{x}_{r}))-\mathbf{x}_{r}\parallel_1\\
&+\mathbb{E}_{\mathbf{x}_s\sim \mathbf{X}_s}\parallel G_{s}(G_{r}(\mathbf{x}_s))-\mathbf{x}_s\parallel_1,
\end{aligned}
\label{equ:cycleConsistency}
\end{equation}
where $G_{s}$, $G_{r}$ are generators from real-to-simulation and simulation-to-real, and $D_{s}$, $D_{r}$ are discriminators for the simulation and real domains, respectively.

Second, based on the binary dropout noise mask $\mathbf{M}=\{0,1\}^{H\times W}$, we can train a pixel-wise rendering network $R$ with the following cross-entropy loss:
\begin{equation}
\small
\begin{aligned}
&\mathcal{L}_{mask}(R,G_s,\mathbf{X}_r,\mathbf{M})=-\mathbb{E}_{(\mathbf{x}_r,\mathbf{m})\sim (\mathbf{X}_r,\mathbf{M})}\\
&\sum_{n=1}^{2}\sum_{h=1}^{H}\sum_{w=1}^{W}\mathds{1}_{[n=\mathbf{m}_{h,w}]}\log(\sigma(R_{n,h,w}(G_s(\mathbf{x}_r)))),
\end{aligned}
\label{equ:mask_loss}
\end{equation}
where $\sigma$ is the softmax function,  $\mathds{1}$ is an indicator function, and $R_{n,h,w}(G_s(\mathbf{x}_r))$ is the value of $F(G_s(\mathbf{x}_r))$ at index $(n,h,w)$. After training $R$, we can render dropout noise for synthetic data and obtain adapted LiDAR images:
\begin{equation}
\small
\mathbf{x}_s^{\prime}=R(\mathbf{x}_s)\odot\mathbf{x}_s,
\label{equ:rendering}
\end{equation}
where $\odot$ is the Hadamard product between a binary mask $R(\mathbf{x}_s)$ and each channel (one matrix) of a tensor $\mathbf{x}_s$.
It is worth noting that SqueezeSegV2 has a Context Aggregation Module (CAM) to eliminate the effect of dropout noise from a feature-level perspective. Differently, our SDNR module predicts a dropout noise map to generate an adapted domain that is similar to the real domain, which can be viewed as a pixel-level dropout noise alignment. Further, these two modules are designed for different purposes: CAM focuses on decreasing the negative effect of dropout noise in one specific domain, while SDNR aims to eliminate the domain gap between two different domains.

\subsection{Statistics-Invariant and Spatially-Adaptive Feature Alignment}
\label{ssec:FeatureAlignment}

\begin{figure}[!t]
\begin{center}
\subfigure{
\includegraphics[width=0.45\linewidth]{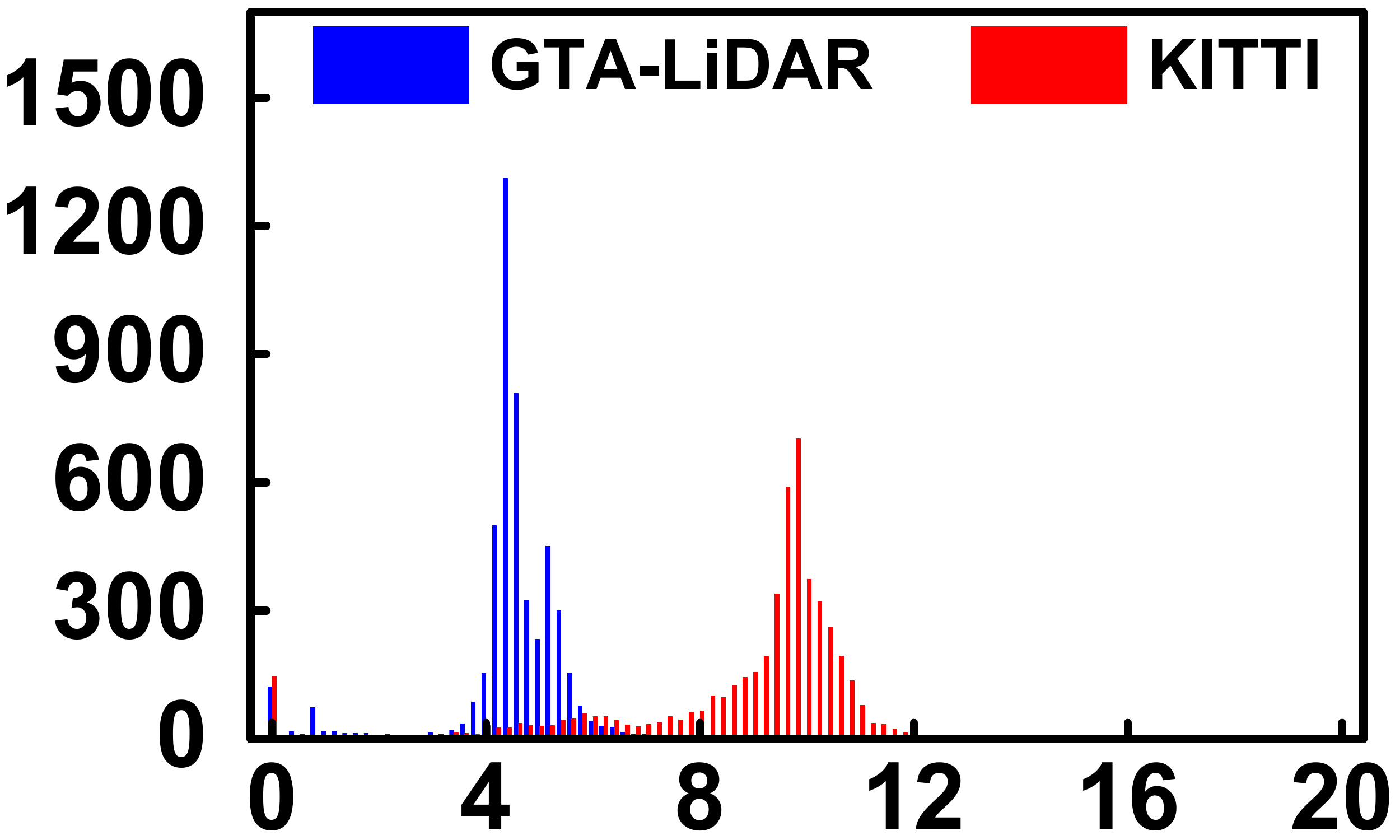}
}
\subfigure{
\includegraphics[width=0.45\linewidth]{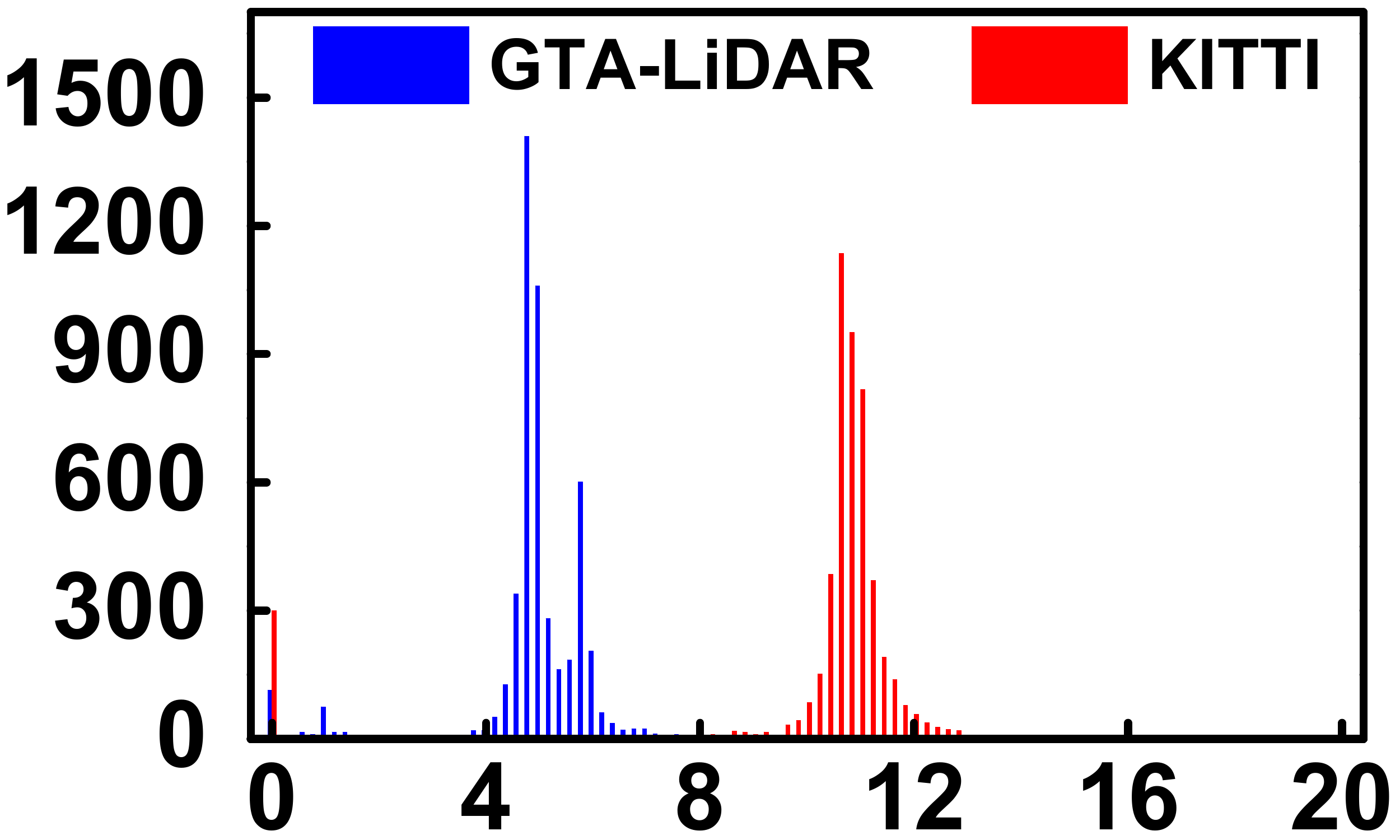}
}
\subfigure{
\includegraphics[width=0.45\linewidth]{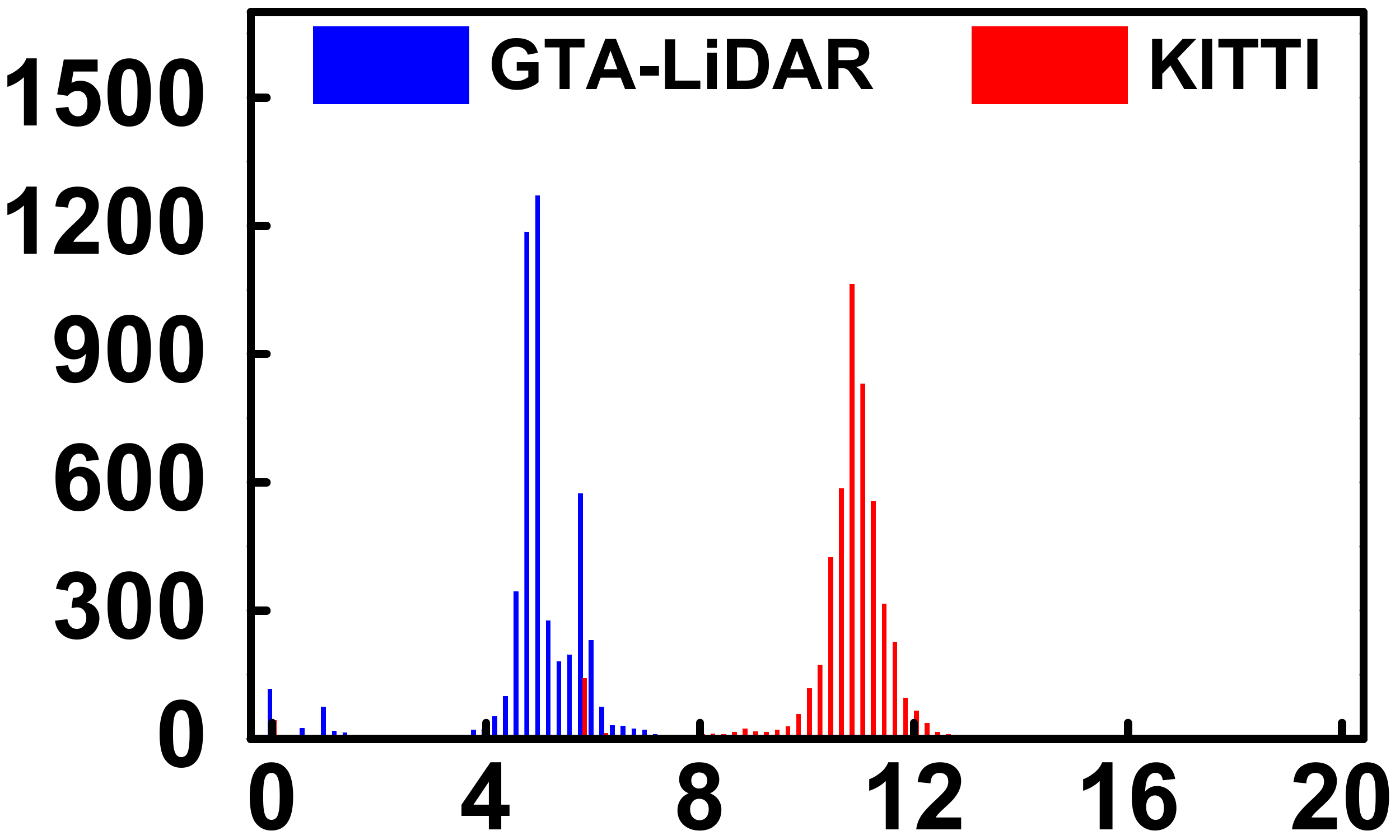}
}
\subfigure{
\includegraphics[width=0.45\linewidth]{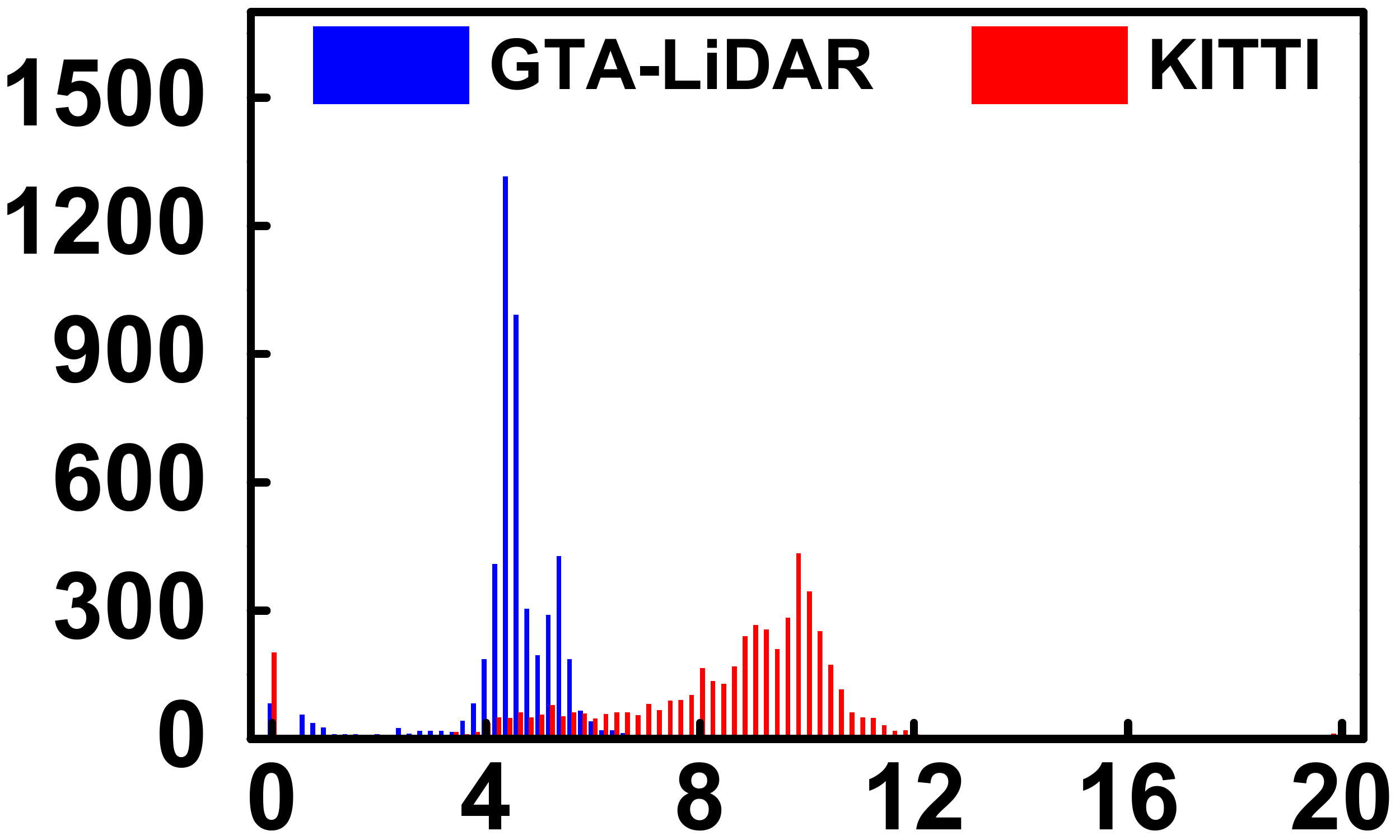}
}
\caption{Pixel-wise feature distribution at four sampled locations on the $x$ coordinate channel of 5,000 projected LiDAR images from GTA-LiDAR~\cite{wu2019squeezesegv2} and KITTI~\cite{geiger2012we,wu2018squeezeseg}. The ordinate represents the number of images, while the abscissa represents the detailed $x$ values.}
\label{fig:SACgap}
\end{center}
\end{figure}

\noindent\textbf{Motivation.}
SqueezeSegV2~\cite{wu2019squeezesegv2} aligns the features between the simulation and real domains during training by Geodesic correlation alignment~\cite{morerio2018minimal}, and employs progressive domain calibration (PDC)~\cite{li2018adaptivebn} to progressively calibrate the statistic shift during post-processing. There are some limitations of this feature alignment method: (1) The PDC module requires the DA pipeline to be designed as multi-stage. Further, it depends heavily on a good sampling of the real-world distribution to obtain accurate statistics, which is very difficult in real applications. (2) The correlation alignment only matches the second-order (covariance) statistics of different distributions, which cannot completely characterize the complex non-Gaussian deep features~\cite{chen2020homm}. 
(3) It neglects the spatial feature gap. \citeauthor{xu2020squeezesegv3}~(\citeyear{xu2020squeezesegv3}) found that the feature distribution of LiDAR images changes drastically at different spatial locations while natural images hold a relatively identical distribution among various locations. Spatially-adaptive convolution (SAC) is proposed by learning a location-wise attention map~\cite{xu2020squeezesegv3}. However, directly transplanting the SAC module into SRDA tasks does not guarantee better performance, because there also exists a spatial feature gap between synthetic LiDAR and real LiDAR, as shown in Figure~\ref{fig:SACgap}.

In this paper, we propose statistic-invariant and spatially-adaptive feature alignment to address the above issues by (1) extracting statistics-invariant features by instance normalization~\cite{ulyanov2016instance}, (2) aligning the feature maps in high-dimensional space by higher-order moment matching~\cite{chen2020homm}, and (3) generating domain-invariant spatial attention map by improving the SAC module~\cite{xu2020squeezesegv3}.

\noindent\textbf{Statistics-Invariant Feature Extraction.} To eliminate the influence of statistics variations among different instances across domains, we employ instance normalization (IN)~\cite{ulyanov2016instance} to normalize each channel of the CNN feature maps, which has been demonstrated to be effective for fast style transfer in RGB images~\cite{wu2018dcan}. Specifically, suppose the feature maps for synthetic image $\mathbf{x}_s^i$ and real image $\mathbf{x}_r^j$ after the same activation layer are $f_{s}^{i}$ and $f_{r}^{f}$ respectively, of the same size $\mathbb{R}^{\hat{C}\times\hat{H}\times\hat{W}}$. We can then easily conduct IN by:
\begin{equation}
\small
\begin{aligned}
&\hat{f}_{s}^{i} = \frac{f_{s}^{i} - \mu(f_{s}^{i})}{\sigma(f_{s}^{i})}, \hat{f}_{r}^{j} = \frac{f_{r}^{j} - \mu(f_{r}^{j})}{\sigma(f_{r}^{j})},\\
&\mu_c(f)=\frac{1}{\hat{H}\hat{W}}\sum\limits^{\hat{H}}_{h=1}\sum\limits^{\hat{W}}_{w=1}f_{chw},\\
& \sigma_c(f)=\sqrt{\frac{1}{\hat{H}\hat{W}}\sum\limits^{\hat{H}}_{h=1}\sum\limits^{\hat{W}}_{w=1}(f_{chw}-\mu_c(f))^2},
\end{aligned}
\label{equ:IN}
\end{equation}
where $\mu_c(f)$ and $\sigma_c(f)$ are the mean and variance across spatial dimensions for the $c$-th channel.

\noindent\textbf{Higher-Order Moment Matching.} We employ higher-order moment matching (HOMM)~\cite{chen2020homm}, a discrepancy-based feature-level alignment method, to align the high-order statistics between the simulation and real domains with the following discrepancy loss:
\begin{equation}
\small
\begin{aligned}
&\mathcal{L}_{HoMM}(\phi,R,\mathbf{X}_{s}, \mathbf{X}_r)=\frac{1}{N^{p}}\\
&\parallel \mathbb{E}_{\mathbf{x}_{s}\sim\mathbf{X}_{s}}(\phi(R(\mathbf{x}_s)\odot\mathbf{x}_s)^{\otimes p})-\mathbb{E}_{\mathbf{x}_{r}\sim\mathbf{X}_{r}}(\phi(\mathbf{x}_{r})^{\otimes p})\parallel^{2}_{F},
\end{aligned}
\label{equ:HOMM}
\end{equation}
where $\parallel\cdot\parallel_F$ is the Frobenius norm, $\phi(\cdot)$ and $N$ denote the activation outputs and the number of hidden neurons of the adapted layer respectively, and $\phi^{\otimes p}$ represents the $p$-level tensor power of the vector $\phi$.

\begin{figure}[!t]
\centering
\includegraphics[width=1.0\linewidth]{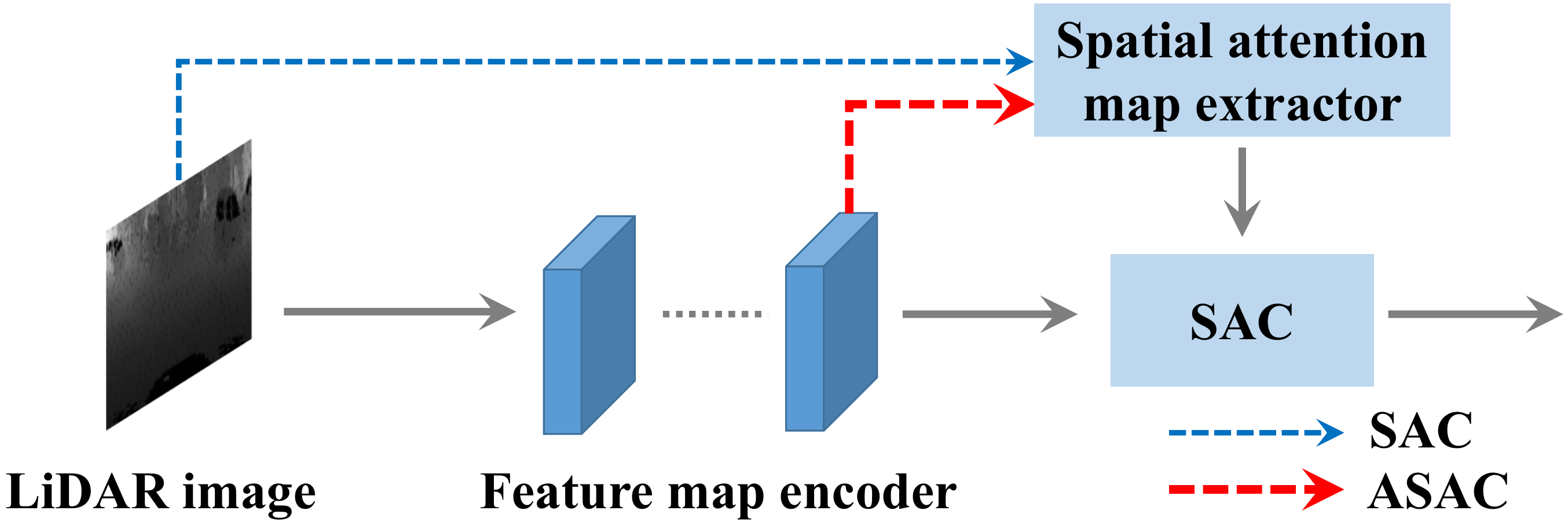}
\caption{The specific pipeline of the ASAC module.}
\label{fig:ASAC}
\end{figure}

\noindent\textbf{Domain-Invariant Spatial Attention Generation.} SAC \cite{xu2020squeezesegv3} introduces one convolution to learn a location-wise attention map. To eliminate the spatial feature gap, we modify the structure by extracting the attention map from the previous feature maps instead of from the original images~\cite{xu2020squeezesegv3}, as shown in Figure~\ref{fig:ASAC}. The basic motivation is that by extracting the spatial attention map according to preceding feature maps, we can align those feature maps between different domains so that the inputs of the aligned SAC (ASAC) module are those domain-invariant features. Once the ASAC module can only see domain-invariant features during the training stage, it is more robust to generate attention map when dealing with the target data.
It is worth noting that we do not need any extra operation because those feature maps have already been aligned by the employed HoMM method.

\subsection{Transferable Segmentation Learning}
\label{ssec:Segmentation}

After generating adapted LiDAR images that have similar styles to real images and aligning the features of the adapted images and real images, we can train a transferable task segmentation model $F$ based on adapted images $\{R(\mathbf{x}_s)\odot\mathbf{x}_s\}$ and corresponding synthetic labels $\mathbf{Y}_s$ with the following focal loss~\cite{lin2017focal,wu2019squeezesegv2}:
\begin{equation}
\small
\begin{aligned}
\mathcal{L}_{seg}(F,R,\mathbf{X}_s,\mathbf{Y}_s)=&-\mathbb{E}_{(\mathbf{x}_s,\mathbf{y}_s)\sim (\mathbf{X}_s,\mathbf{Y}_s)}\sum\limits_{l=1}^{L}\sum\limits_{h=1}^{H}\sum\limits_{w=1}^{W}
\\&\mathds{1}_{[l=\mathbf{y}_{s(h,w)}]}
(1-p_{l,h,w})^\gamma\log p_{l,h,w},
\end{aligned}
\label{equ:seg_loss}
\end{equation}
where $p_{l,h,w}=\sigma(F_{l,h,w}(R(\mathbf{x}_s)\odot\mathbf{x}_s))$, $F_{l,h,w}(\cdot)$ is the value of $F(\cdot)$ at index $(l,h,w)$, and $\gamma$ is a focusing parameter to adjust the rate at which well-classified examples are down-weighted. When $\gamma = 0$, focal loss equals cross-entropy loss. The advantage of focal loss is that it can deal with the imbalanced distribution of point cloud categories, \textit{e.g.} more background points than foreground objects.

\subsection{ePointDA Learning}
\label{ssec:Learning}

The proposed ePointDA learning framework utilizes adaptation techniques to bridge the domain shift at both pixel-level and feature-level. Combining these alignment modules with transferable segmentation learning, we can obtain the overall objective loss function of ePointDA as:
\begin{equation}
\small
\begin{aligned}
&\mathcal{L}_{ePointDA}(G_s,D_s,G_r,D_r,R,F)=\\
&\mathcal{L}_{GAN}^{r\rightarrow s}(G_{s},D_s, \mathbf{X}_r,\mathbf{X}_s)+\mathcal{L}_{GAN}^{s\rightarrow r}(G_{r},D_r,\mathbf{X}_s,\mathbf{X}_r)+\\
&\mathcal{L}_{cyc}(G_{s},G_{r},\mathbf{X}_{r}, \mathbf{X}_s)+\mathcal{L}_{mask}(R,G_s,\mathbf{X}_r,\mathbf{M})+\\
&\mathcal{L}_{HoMM}(F_{\phi},R,\mathbf{X}_{s}, \mathbf{X}_r)+\mathcal{L}_{seg}(F,R,\mathbf{X}_s,\mathbf{Y}_s),
\end{aligned}
\label{equ:total_loss}
\end{equation}
where $F_{\phi}$ is the activation outputs of $F$ used for HoMM. The training process corresponds to solving for the target segmentation model $F$ according to:
\begin{equation}
\small
\begin{aligned}
F^*=&\arg\min_F\min_R\min_{D_s,D_r}\max_{G_s,G_r}\\
&\mathcal{L}_{ePointDA}(G_s,D_s,G_r,D_r,R,F).
\end{aligned}
\label{equ:optimization}
\end{equation}
As shown in Eq.~(\ref{equ:total_loss}), there are 6 different losses in the overall objective loss function. The SDNR module involves 4 losses: GAN losses in two directions, cycle-consistency loss, and dropout noise rendering (DNR) loss. The feature alignment and segmentation learning modules correspond to HoMM loss and segmentation supervision loss, respectively. The GAN losses and cycle-consistency loss are used to update the CycleGAN network while the DNR loss is used to update the dropout noise prediction network. The gradient backpropagation of DNR loss would be truncated at the position of dropout noise prediction network’s inputs. Similarly, the HoMM loss and segmentation supervision loss are used to update the segmentation network, indicating that the gradient will be backpropagated only to the segmentation network.

\begin{table}[!t]
\centering\footnotesize
\resizebox{\linewidth}{!}{%
\begin{tabular}{c | c  c  c  |  c  c  c }
\toprule
 \multirow{2}{*}{\centering Method} & \multicolumn{3}{c|}{\centering Car} & \multicolumn{3}{c}{\centering Pedestrian}\\
\cline{2-7}
 & Pre & Rec & IoU & Pre & Rec & IoU \\
\hline
 Source-only & 34.4 & 67.6 & 29.6 & 11.3 & 8.9 & 5.2 \\
\hline
 DAN~\cite{long2015learning} & 56.3 & 76.4 & 47.8 & 20.8 & 68.9 & 19.0 \\
 CORAL~\cite{dcoral} & 56.5 & 82.1 & 50.2 & 26.0 & 50.3 & 20.7 \\
 HoMM~\cite{chen2020homm} & 59.4 & 85.2 & 53.9 & 26.2 & 66.8 & 23.2 \\
\hline
 ADDA~\cite{tzeng2017adversarial} & 56.7 & 83.5 & 50.7 & 24.7 & 58.5 & 21.0 \\
\hline
 CyCADA~\cite{hoffman2018CyCADA} & 40.9 & 72.1 & 35.3 & 17.8 & 52.4 & 15.3 \\
\hline
 SqueezeSegV2~\cite{wu2019squeezesegv2} & - & - & 57.4 & - & - & 23.5 \\
\hline
 \textbf{ePointDA (Ours)} & 75.2 & 84.7 & \textbf{66.2} & 28.7 & 65.2 & \textbf{24.8} \\
\hline
 Oracle~(SqueezeSegV2) & 76.7 & 92.1 & 71.9 & 28.5 & 82.3 & 26.9 \\	
 Oracle+SAC & 78.4 & 91.4 & 73.1 & 28.9 & 84.5 & 27.4 \\
 Oracle+SAC+HHead & 77.8 & 93.1 & 73.5 & 29.3 & 86.6 & 28.0 \\	
\bottomrule
\end{tabular}
}
\caption{Comparison with the state-of-the-art DA methods for LiDAR point cloud segmentation from GTA-LiDAR to KITTI, where +ASAC denotes using the  spatial feature aligned SAC module, and +HHead denotes replacing the CRF layer with an conv layer. The best IoU of each category trained on the simulation domain is emphasized in bold.}
\label{tab:GTA_KITTI}
\end{table}

\section{Experiments}
In this section, we introduce the experimental settings and compare ePointDA with state-of-the-art approaches, followed by series of ablation studies and visualizations.

\subsection{Experimental Settings}

\noindent\textbf{Datasets.} 
We perform SRDA from synthetic GTA-LiDAR~\cite{wu2019squeezesegv2} to real KITTI~\cite{geiger2012we} and SemanticKITTI~\cite{behley2019semantickitti} datasets for LiDAR point cloud segmentation. GTA-LiDAR~\cite{wu2019squeezesegv2} contains 100,000 LiDAR point clouds synthesized in GTA-V. The depth segmentation map is generated by~\cite{krahenbuhl2018free} and the Image-LiDAR registration is conducted by~\cite{yue2018LiDAR}. There are one label and $x, y, z$ coordinates for each point in the synthetic point cloud, without dropout noise and intensity.

KITTI~\cite{geiger2012we,wu2018squeezeseg} contains 10,848 samples with point-wise labels obtained from the original 3D bounding boxes. As in SqueezeSegV2~\cite{wu2019squeezesegv2}, we split the dataset into a training set with 8,057 samples and a test set with 2,791 samples.

SemanticKITTI~\cite{behley2019semantickitti} is a recently released large-scale dataset for LiDAR point-cloud segmentation with 21 sequences and 43,442 densely annotated scans. Following~\cite{behley2019semantickitti}, we employ sequences-\{0-7\} and \{9, 10\} (19,130 scans) for training, sequence-08 (4,071 scans) for validation, and sequences-\{11-21\} (20,351 scans) for testing.

Since there are only two categories in the GTA-LiDAR dataset, \textit{i.e.} car and pedestrian, we select the images in KITTI and SemanticKITTI that contain these two categories\footnote{For SemanticKITTI, we combine the original \{car, truck, bus\} and \{person\} respectively as the car and pedestrian categories.} and report the segmentation adaptation results. Constructing a synthetic dataset with more categories for LiDAR point cloud segmentation and performing SRDA remains our future work.

\begin{table}[!t]
\centering\footnotesize
\resizebox{\linewidth}{!}{%
\begin{tabular}{c | c  c  c  |  c  c  c }
\toprule
 \multirow{2}{*}{\centering Method} & \multicolumn{3}{c|}{\centering Car} & \multicolumn{3}{c}{\centering Pedestrian}\\
\cline{2-7}
 & Pre & Rec & IoU & Pre & Rec & IoU \\
\hline
 Source-only & 63.8 & 40.8 & 33.2 & 3.9 & 43.7 & 3.7 \\
\hline
 DAN~\cite{long2015learning} & 73.6 & 66.4 & 53.6 & 12.1 & 28.0 & 9.2 \\
 CORAL~\cite{dcoral} & 73.4 & 71.3 & 56.6 & 8.9 & 30.0 & 7.4 \\
 HoMM~\cite{chen2020homm} & 73.8 & 69.5 & 55.7 & 11.7 & 23.8 & 8.5 \\
\hline
 ADDA~\cite{tzeng2017adversarial} & 65.3 & 84.5 & 58.3 & 13.7 & 30.1 & 10.4 \\
\hline
 CyCADA~\cite{hoffman2018CyCADA} & 78.3 & 48.1 & 42.4 & 9.7 & 34.8 & 8.2 \\
\hline
 SqueezeSegV2~\cite{wu2019squeezesegv2} & 65.9 & 93.8 & 63.2 & 14.9 & 46.9 & 12.8 \\
\hline
 \textbf{ePointDA (Ours)} & 77.9 & 88.5 & \textbf{70.7} & 14.7 & 51.2 & \textbf{12.9} \\
\hline
 Oracle~(SqueezeSegV2) & 93.9 & 96.4 & 90.7 & 41.4 & 50.0 & 29.3 \\	
 Oracle+SAC & 92.9 & 98.2 & 91.4 & 47.0 & 46.4 & 30.4 \\
 Oracle+SAC+HHead & 93.4 & 98.5 & 92.1 & 49.3 & 48.8 & 32.5 \\	
\bottomrule
\end{tabular}
}
\caption{Comparison with the state-of-the-art DA methods from GTA-LiDAR to SemanticKITTI.}
\label{tab:GTA_SemanticKITTI}
\end{table}

\noindent\textbf{Evaluation Metrics.}
Similar to~\cite{wu2018squeezeseg}, we employ precision, recall, and intersection-over-union (IoU) to evaluate the class-level segmentation results by comparing the predicted results with ground-truth labels point-wisely:
$Pre_l = \frac{|\mathcal{P}_l \cap \mathcal{G}_l|}{|\mathcal{P}_l},Rec_l = \frac{|\mathcal{P}_l \cap \mathcal{G}_l|}{|\mathcal{G}_l},IoU_l = \frac{|\mathcal{P}_l \cap \mathcal{G}_l|}{|\mathcal{P}_l \cup \mathcal{G}_l|},$
where $\mathcal{P}_l$ and $\mathcal{G}_l$ respectively denote the predicted and ground-truth point sets that belong to class-$l$, and $|\cdot|$ represents the cardinality of a set. Larger precision, recall, and IoU values represent better results. We employ IoU as the primary metric.

\noindent\textbf{Baselines.}
To the best of our knowledge, ePointDA is the first end-to-end framework for simulation-to-real LiDAR point cloud segmentation. To demonstrate its effectiveness, we compare to baselines of three types: (1) source-only, directly transferring the model trained on the simulation domain; (2) SqueezeSegV2~\cite{wu2019squeezesegv2}, one state-of-the-art SRDA method for LiDAR point cloud segmentation; (3) state-of-the-art DA methods for RGB image classification and segmentation tasks: DAN~\cite{long2015learning}, CORAL~\cite{dcoral}, ADDA~\cite{tzeng2017adversarial}, CyCADA~\cite{hoffman2018CyCADA}, and HoMM~\cite{chen2020homm}. For comparison, we also report the oracle result, \textit{i.e.} the segmentation model is trained also on the real domain, which can be viewed as an upper bound.

\noindent\textbf{Implementation Details.}
For SDNR, we employ the same architecture as SqueezeSegV2~\cite{wu2019squeezesegv2} except the final layer which is changed to be a binary mask classification. The segmentation network is shown in Figure~\ref{fig:segmentationNetwork}, which replaces all the standard convolution and batch normalization in SqueezeSegV2~\cite{wu2019squeezesegv2} with spatially-adaptive convolution and instance normalization. Our model is implemented using TensorFlow with the same hyper-parameters as~\cite{wu2019squeezesegv2}. We use stochastic gradient descent (SGD) as the optimizer with a momentum of 0.9 and a batch size of 20. The initial learning rate is 0.05 with a decay factor of 0.5 for every 20,000 steps. As in~\cite{lin2017focal,wu2019squeezesegv2}, the focusing parameter $\gamma$ in Eq.~(\ref{equ:seg_loss}) is set to $2$. Besides, in practical implementation, the complexity of calculating the higher-order tensor $\phi^{\otimes p}$ in Eq.~(\ref{equ:HOMM}) would explode (\textit{i.e.} $\mathcal{O}(L^{p})$) as order $p$ increases. Following~\cite{chen2020homm}, we use
the Monte Carlo estimation as an approximation.

\begin{table}[!t]
\centering\footnotesize
\resizebox{\linewidth}{!}{%
\begin{tabular}{c | c  c  c  |  c  c  c }
\toprule
 \multirow{2}{*}{\centering Method} & \multicolumn{3}{c|}{\centering Car} & \multicolumn{3}{c}{\centering Pedestrian}\\
\cline{2-7}
 & Pre & Rec & IoU & Pre & Rec & IoU \\
\hline
 Baseline & 59.6 & 83.2 & 53.1 & 21.5 & 77.1 & 20.2 \\
 +SDNR & 65.7 & 84.8 & 58.7 & 24.7 & 75.4 & 22.8 \\
 +SDNR+IN & 69.3 & 86.9 & 62.7 & 28.8 & 57.6 & 23.8 \\
 +SDNR+IN+HoMM & 73.4 & 81.9 & 63.4 & 29.4 & 56.0 & 23.9  \\
 +SDNR+IN+HoMM+ASAC & 72.1 & 85.6 & 64.2 & 31.2 & 57.5 & \textbf{25.3}  \\
 +SDNR+IN+HoMM+ASAC+HHead & 75.2 & 84.7 & \textbf{66.2} & 28.7 & 65.2 & 24.8 \\
\bottomrule
\end{tabular}
}
\caption{Ablation study on different modules, where Baseline denotes a simplified SqueezeSegV2 model~\cite{wu2019squeezesegv2} for fair comparison taking the Cartesian coordinates as input and using batch normalization, frequency-based DNR, and Geodesic correlation alignment.}
\label{tab:Ablation}
\end{table}

\subsection{Comparison with the State-of-the-art}
\label{ssec:comparisonSOTA}

The performance comparisons between ePointDA and the state-of-the-art DA methods are shown in Table~\ref{tab:GTA_KITTI} and Table~\ref{tab:GTA_SemanticKITTI}. From the results, we can observe that\footnote{Since pedestrian is synthesized by a very simple physical model in the GTA-LiDAR dataset~\cite{wu2019squeezesegv2}, which often represents pedestrians as cylinders, we mainly focus on the IoU of the ``car'' class for evaluation and comparison.}:

(1) Because of the presence of domain shift, the joint probability distributions of observed LiDAR point clouds and corresponding labels significantly vary between the simulation and real domains. As a result, the source-only method that does not consider the domain gap obtains the worst IoU performance, which motivates the research on domain adaptation.


(2) On one hand, when directly applied to the LiDAR SRDA task, the traditional RGB-image-targeted DA methods outperform the source-only setting. Among these methods, discrepancy-based methods (DAN, CORAL, HoMM) and adversarial discriminative methods (ADDA) obtain much better results than adversarial generative methods (CyCADA). Besides feature-level alignment, CyCADA also conducts pixel-level alignment by CycleGAN~\cite{zhu2017unpaired}, which is mainly designed to translate natural images' colors and textures among different domains. Since the projected 2D LiDAR images are mainly about geometric information, which is quite different from RGB images, the translated LiDAR images by CycleGAN are of low quality. On the other hand, these RGB-image-targeted DA methods do not outperform SqueezeSegV2~\cite{wu2019squeezesegv2}, one SRDA method specially designed for LiDAR point cloud segmentation. 
This is reasonable, because they do not consider the specific characteristics of LiDAR point clouds, such as dropout noise, the intensity of synthetic data, and the statistics difference between simulation and real domains.

(3) ePointDA performs the best among all adaptation settings. For example, the performance improvements of ePointDA over SqueezeSegV2 are 8.8\% and 7.5\% on KITTI and SemanticKITTI, respectively. These results demonstrate the superiority of ePointDA relative to state-of-the-art DA approaches for LiDAR point cloud segmentation\footnote{One may argue that ePointDA contains some extra modules (\textit{i.e.} Aligned SAC and one more conv layer), which might be unfair to compare with SqueeseSegV2. Even if we drop these modules, \textit{i.e.} +SDNR+IN+HoMM in Table~\ref{tab:Ablation}, our method still outperforms SqueeseSegV2 by a large margin (63.4 vs. 57.4).}. The performance improvements benefit from the advantages of ePointDA: pixel-level alignment by self-supervised dropout noise rendering and feature-level alignment by higher-order moment matching with statistics-invariant features and domain-invariant spatial attentions.

(4) There is still a large performance gap between ePointDA and the oracle method. As demonstrated in~\cite{xu2020squeezesegv3}, the SAC module can improve the oracle performance. Replacing the CRF layer in SqueezeSegV2 with a conv layer also works under the oracle setting.

\begin{table}[!t]
\centering\footnotesize
\resizebox{\linewidth}{!}{%
\begin{tabular}{c | c | c  c  c  |  c  c  c }
\toprule
 \multirow{2}{*}{\centering DNR} & \multirow{2}{*}{\centering Norm} & \multicolumn{3}{c|}{\centering Car} & \multicolumn{3}{c}{\centering Pedestrian}\\
\cline{3-8}
 & & Pre & Rec & IoU & Pre & Rec & IoU \\
\hline
 \multirow{4}{*}{\centering Frequency~\cite{wu2019squeezesegv2}} & BN & 59.6 & 83.2 & 53.1 & 21.5 & 77.1 & 20.2 \\
 & IN & 63.5 & 83.6 & \textbf{56.5} & 25.4 & 65.6 & \textbf{22.4} \\
 & LN & 60.2 & 84.6 & 54.3 & 24.3 & 65.5 & 21.5\\
 & GN & 61.4 & 83.5 & 54.7 & 26.0 & 50.3 & 20.7 \\
\hline
  \multirow{4}{*}{\centering Learned (ours)} & BN &  65.7 & 84.8 & 58.7 & 24.7 & 75.4 & 22.8 \\
 & IN & 69.3 & 86.9 & \textbf{62.7} & 28.8 & 57.6 & \textbf{23.8} \\
 & LN & 65.1 & 88.6 & 60.1 & 27.3 & 59.3 & 22.9 \\
 & GN  & 64.1 & 88.9 & 59.3 & 27.8 & 61.4 & 23.7 \\
\bottomrule
\end{tabular}
}
\caption{Ablation study on different normalization schemes using both frequency-based DNR and our learned DNR without feature alignment. `BN', `IN', `LN', `GN' are short for batch normalization, instance normalization, layer normalization, and group normalization, respectively.}
\label{tab:Ablation_Normalization}
\end{table}

\begin{figure*}
\centering
\includegraphics[width=1.0\linewidth]{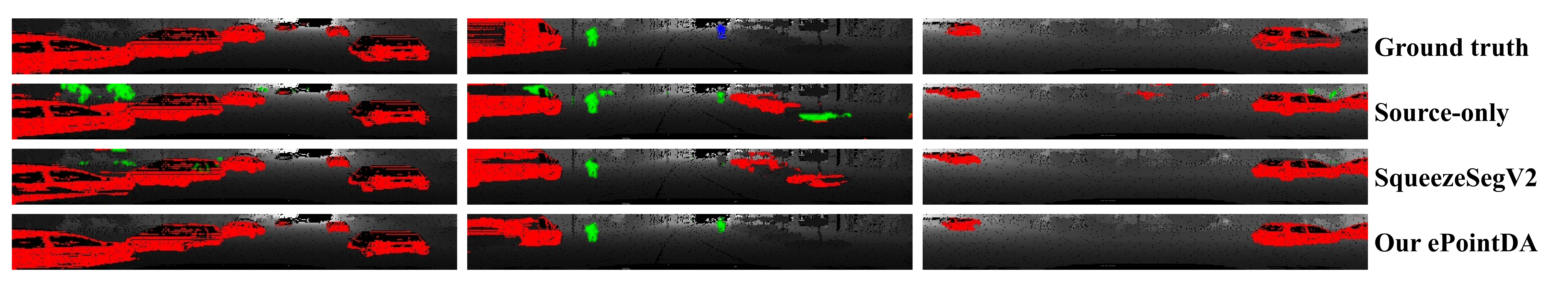}
\caption{Qualitative segmentation result from synthetic GTA-LiDAR to real KITTI (red: car, green: pedestrian, blue: cyclist).}
\label{fig:SegmentationResult}
\end{figure*}

\begin{figure*}
\centering
\includegraphics[width=1.0\linewidth]{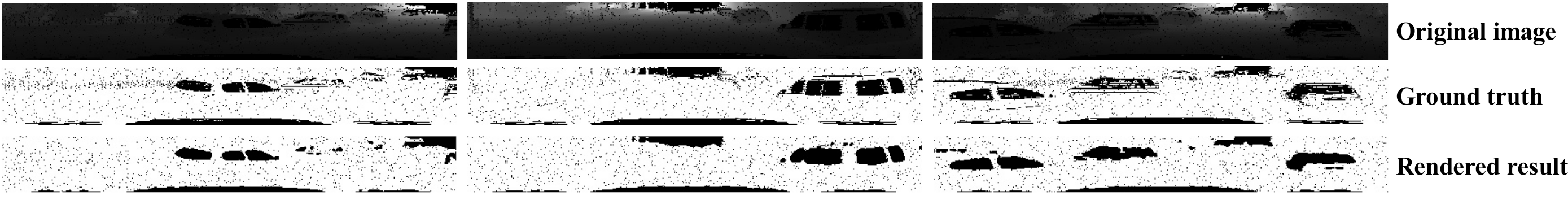}
\caption{Rendered vs. ground truth dropout noise on the real KITTI dataset.}
\label{fig:DropoutNoiseRendering}
\end{figure*}

\subsection{Ablation Study}

We conduct a series of ablation studies when adapting from GTA-LiDAR to KITTI. First, we incrementally investigate the effectiveness of different modules in ePointDA. From the results in Table~\ref{tab:Ablation}, we can observe that: (1) adding each module can improve the IoU scores, which demonstrates that all the modules contained in ePointDA contribute to the SRDA task; (2) among all these modules, SDNR provides the highest performance improvement (5.6\%), which demonstrates the important role that dropout noise plays in the domain gap between the simulation and real domains and the necessity of exploring effective DNR model.

Second, we explore the differences caused by various normalization schemes, including batch normalization (BN)~\cite{ioffe2015batch}, instance normalization (IN)~\cite{ulyanov2016instance}, layer normalization (LN)~\cite{ba2016layer}, and group normalization (GN)~\cite{wu2018group}. From Table~\ref{tab:Ablation_Normalization}, it is clear that IN, LN, and GN all outperform BN. The relative poor performance of BN results from the statistics gap between the simulation and real domains. IN, LN, and GN can all eliminate such gap to some extent~\cite{wu2018dcan} and thus achieve better DA results than BN.

\begin{table}[!t]
\centering\footnotesize
\resizebox{\linewidth}{!}{%
\begin{tabular}{c |  c  c  c  |  c  c  c }
\toprule
 \multirow{2}{*}{\centering Method} & \multicolumn{3}{c|}{\centering Car} & \multicolumn{3}{c}{\centering Pedestrian}\\
\cline{2-7}
 & Pre & Rec & IoU & Pre & Rec & IoU \\
\hline
 Baseline & 73.4 & 81.9 & 63.4 & 29.4 & 56.0 & 23.9 \\
 SAC~\cite{xu2020squeezesegv3} & 68.5 & 83.2 & 60.2 & 25.2 & 62.4 & 21.9 \\
 ASAC (ours) & 72.1 & 85.6 & \textbf{64.2} & 31.2 & 57.5 & \textbf{25.3} \\
\bottomrule
\end{tabular}
}
\caption{Comparison between ordinary SAC~\cite{xu2020squeezesegv3} and our aligned SAC (ASAC). Baseline corresponds the ``+SDNR+IN+HoMM'' setting in Table~\ref{tab:Ablation}.}
\label{tab:Ablation_SAC}
\end{table}

Third, we compare our aligned SAC (ASAC) and the ordinary SAC~\cite{xu2020squeezesegv3} in Table~\ref{tab:Ablation_SAC}. Without considering the spatial feature gap, simply incorporating the ordinary SAC can result in a significant performance drop. Our aligned modification helps address this issue and thus improves the DA performance.

Finally, we study the influence of convolution (conv) layers appended to the last deconvolution layer of the segmentation network. From Table~\ref{tab:Ablation_convLayers}, we can conclude that adding 2 conv layers performs the best. As stated in~\cite{li2019scale, Yu2016, dai2017deformable, dai2016r}, the receptive field can affect the network's effectiveness. As the number of conv layers increases, the segmentation results become better; after reaching the best receptive field (2 conv layers), the segmentation results decrease gradually.

\begin{table}[!t]
\centering\small
\begin{tabular}{c | c  c  c  |  c  c  c }
\toprule
 \multirow{2}{*}{\centering \#Conv} & \multicolumn{3}{c|}{\centering Car} & \multicolumn{3}{c}{\centering Pedestrian}\\
\cline{2-7}
 & Pre & Rec & IoU & Pre & Rec & IoU \\
\hline
 1 & 72.1 & 85.6 & 64.2 & 31.2 & 57.5 & \textbf{25.3} \\
 2 & 75.2 & 84.7 & \textbf{66.2} & 28.7 & 65.2 & 24.8  \\
 3 & 74.1 & 82.3 & 63.8 & 26.9 & 64.8 & 23.4 \\
 4 & 70.6 & 83.4 & 61.9 & 25.1 & 59.7 & 21.5 \\			
 5 & 68.6 & 84.0 & 60.7 & 26.5 & 52.3 & 21.3 \\
\bottomrule
\end{tabular}
\caption{Ablation study on the number of convolution layers (\#Conv) that are appended to the last deconvolution layer. This experiment is conducted after dropout noise rendering and feature alignment, \textit{i.e.} +SDNR+IN+HoMM+ASAC.
}
\label{tab:Ablation_convLayers}
\end{table}

\subsection{Visualization}
First, we qualitatively visualize the LiDAR point cloud segmentation results from GTA-LiDAR to KITTI in Figure~\ref{fig:SegmentationResult}. We can clearly see that after adaptation by ePointDA, the segmentation results are improved notably as compared to source-only and SqueezeSegV2~\cite{wu2019squeezesegv2}. For example, in the first column, ePointDA avoids falsely detecting some pedestrians. In the second column, ePointDA classifies the cyclist as a pedestrian, which is reasonable because there is no cyclist in the GTA-LiDAR dataset. Please note that there does exist some missing objects in the original annotations, such as the car on the right corner in the third column. This is either because some objects are not labeled in the original 3D point clouds, or because some information is missing during the projection from 3D point clouds to 2D images. For fair comparison, we used the same setting as SqueezeSegV2. We will refine the annotations in our future work to make the proposed method more practical.

Second, we visualize the DNR results on KITTI. From the results in Figure~\ref{fig:DropoutNoiseRendering}, it is clear that the rendered dropout noise is very close to the ground truth, which demonstrates the effectiveness of our SDNR method.

\section{Conclusion}
In this paper, we proposed an end-to-end simulation-to-real domain adaptation (SRDA) framework, named ePointDA, for LiDAR point cloud segmentation. By explicitly rendering dropout noise for the real domain in a self-supervised manner and spatially aligning higher-level moments between the simulation and real domains, ePointDA bridges the domain shift at both the pixel-level and feature-level. Further, ePointDA does not require prior statistics of the real domain, which makes it more robust and practical. The extensive experiments adapting from synthetic GTA-LiDAR to real KITTI and SemanticKITTI demonstrated that ePointDA significantly outperforms the state-of-the-art SRDA methods. The proposed ePointDA can be easily applied to other applications, such as robotics grasping. It can also be applied to other depth data, like the depth from Kinect, which has similar properties to LiDAR. Another promising extension is the specific adaptation setting where the source and target domains have different channels, such as adapting from RGB to RGB-Depth. 

In future studies, we plan to construct a large-scale synthetic dataset for LiDAR point cloud segmentation containing more compatible categories with SemanticKITTI and extend our framework to corresponding SRDA tasks. We will explore multi-modal domain adaptation by jointly modeling multiple modalities, such as image and LiDAR.

\section{Acknowledgments}
This work is supported by the National Natural Science Foundation of China (No. 61701273), Berkeley DeepDrive, and the Zhongguancun Haihua Institute for Frontier Information Technology.

\small\bibliography{ePointDA}

\begin{thebibliography}{78}
\providecommand{\natexlab}[1]{#1}
\providecommand{\url}[1]{\texttt{#1}}
\providecommand{\urlprefix}{URL }
\expandafter\ifx\csname urlstyle\endcsname\relax
  \providecommand{\doi}[1]{doi:\discretionary{}{}{}#1}\else
  \providecommand{\doi}{doi:\discretionary{}{}{}\begingroup
  \urlstyle{rm}\Url}\fi

\bibitem[{Achituve, Maron, and Chechik(2021)}]{achituve2021self}
Achituve, I.; Maron, H.; and Chechik, G. 2021.
\newblock Self-Supervised Learning for Domain Adaptation on Point Clouds.
\newblock In \emph{WACV}, 123--133.

\bibitem[{Ba, Kiros, and Hinton(2016)}]{ba2016layer}
Ba, J.~L.; Kiros, J.~R.; and Hinton, G.~E. 2016.
\newblock Layer normalization.
\newblock arXiv:1607.06450v1 [stat.ML].

\bibitem[{Behley et~al.(2019)Behley, Garbade, Milioto, Quenzel, Behnke,
  Stachniss, and Gall}]{behley2019semantickitti}
Behley, J.; Garbade, M.; Milioto, A.; Quenzel, J.; Behnke, S.; Stachniss, C.;
  and Gall, J. 2019.
\newblock SemanticKITTI: A dataset for semantic scene understanding of lidar
  sequences.
\newblock In \emph{ICCV}, 9297--9307.

\bibitem[{Caltagirone et~al.(2017)Caltagirone, Scheidegger, Svensson, and
  Wahde}]{caltagirone2017fast}
Caltagirone, L.; Scheidegger, S.; Svensson, L.; and Wahde, M. 2017.
\newblock Fast LIDAR-based road detection using fully convolutional neural
  networks.
\newblock In \emph{IV}, 1019--1024.

\bibitem[{Carlucci et~al.(2019)Carlucci, D'Innocente, Bucci, Caputo, and
  Tommasi}]{carlucci2019domain}
Carlucci, F.~M.; D'Innocente, A.; Bucci, S.; Caputo, B.; and Tommasi, T. 2019.
\newblock Domain generalization by solving jigsaw puzzles.
\newblock In \emph{CVPR}, 2229--2238.

\bibitem[{Chen et~al.(2020)Chen, Fu, Chen, Jin, Cheng, Jin, and
  Hua}]{chen2020homm}
Chen, C.; Fu, Z.; Chen, Z.; Jin, S.; Cheng, Z.; Jin, X.; and Hua, X.-S. 2020.
\newblock HoMM: Higher-order Moment Matching for Unsupervised Domain
  Adaptation.
\newblock In \emph{AAAI}, 3422--3429.

\bibitem[{Dai et~al.(2016)Dai, Li, He, and Sun}]{dai2016r}
Dai, J.; Li, Y.; He, K.; and Sun, J. 2016.
\newblock R-fcn: Object detection via region-based fully convolutional
  networks.
\newblock In \emph{NeurIPS}, 379--387.

\bibitem[{Dai et~al.(2017)Dai, Qi, Xiong, Li, Zhang, Hu, and
  Wei}]{dai2017deformable}
Dai, J.; Qi, H.; Xiong, Y.; Li, Y.; Zhang, G.; Hu, H.; and Wei, Y. 2017.
\newblock Deformable convolutional networks.
\newblock In \emph{ICCV}, 764--773.

\bibitem[{Dosovitskiy et~al.(2017)Dosovitskiy, Ros, Codevilla, Lopez, and
  Koltun}]{dosovitskiy2017carla}
Dosovitskiy, A.; Ros, G.; Codevilla, F.; Lopez, A.; and Koltun, V. 2017.
\newblock CARLA: An open urban driving simulator.
\newblock In \emph{CoRL}, 1--16.

\bibitem[{Dovrat, Lang, and Avidan(2019)}]{dovrat2019learning}
Dovrat, O.; Lang, I.; and Avidan, S. 2019.
\newblock Learning to sample.
\newblock In \emph{CVPR}, 2760--2769.

\bibitem[{Feng, Xu, and Tao(2019)}]{feng2019self}
Feng, Z.; Xu, C.; and Tao, D. 2019.
\newblock Self-Supervised Representation Learning From Multi-Domain Data.
\newblock In \emph{ICCV}, 3245--3255.

\bibitem[{Geiger, Lenz, and Urtasun(2012)}]{geiger2012we}
Geiger, A.; Lenz, P.; and Urtasun, R. 2012.
\newblock Are we ready for autonomous driving? the kitti vision benchmark
  suite.
\newblock In \emph{CVPR}, 3354--3361.

\bibitem[{Goodfellow et~al.(2014)Goodfellow, Pouget-Abadie, Mirza, Xu,
  Warde-Farley, Ozair, Courville, and Bengio}]{goodfellow2014generative}
Goodfellow, I.; Pouget-Abadie, J.; Mirza, M.; Xu, B.; Warde-Farley, D.; Ozair,
  S.; Courville, A.; and Bengio, Y. 2014.
\newblock Generative adversarial nets.
\newblock In \emph{NeurIPS}, 2672--2680.

\bibitem[{Hoffman et~al.(2018)Hoffman, Tzeng, Park, Zhu, Isola, Saenko, Efros,
  and Darrell}]{hoffman2018CyCADA}
Hoffman, J.; Tzeng, E.; Park, T.; Zhu, J.-Y.; Isola, P.; Saenko, K.; Efros,
  A.~A.; and Darrell, T. 2018.
\newblock CyCADA: Cycle-Consistent Adversarial Domain Adaptation.
\newblock In \emph{ICML}, 1994--2003.

\bibitem[{Huang, Wang, and Neumann(2018)}]{huang2018recurrent}
Huang, Q.; Wang, W.; and Neumann, U. 2018.
\newblock Recurrent slice networks for 3d segmentation of point clouds.
\newblock In \emph{CVPR}, 2626--2635.

\bibitem[{Ioffe and Szegedy(2015)}]{ioffe2015batch}
Ioffe, S.; and Szegedy, C. 2015.
\newblock Batch Normalization: Accelerating Deep Network Training by Reducing
  Internal Covariate Shift.
\newblock In \emph{ICML}, 448--456.

\bibitem[{Jiang et~al.(2019)Jiang, Zhao, Liu, Shen, Fu, and
  Jia}]{jiang2019hierarchical}
Jiang, L.; Zhao, H.; Liu, S.; Shen, X.; Fu, C.-W.; and Jia, J. 2019.
\newblock Hierarchical point-edge interaction network for point cloud semantic
  segmentation.
\newblock In \emph{ICCV}, 10433--10441.

\bibitem[{Jiang and Saripalli(2020)}]{jiang2020LiDARnet}
Jiang, P.; and Saripalli, S. 2020.
\newblock LiDARNet: A Boundary-Aware Domain Adaptation Model for Lidar Point
  Cloud Semantic Segmentation.
\newblock arXiv:2003.01174v2 [cs.CV].

\bibitem[{Klokov and Lempitsky(2017)}]{klokov2017escape}
Klokov, R.; and Lempitsky, V. 2017.
\newblock Escape from cells: Deep kd-networks for the recognition of 3d point
  cloud models.
\newblock In \emph{ICCV}, 863--872.

\bibitem[{Kr{\"a}henb{\"u}hl(2018)}]{krahenbuhl2018free}
Kr{\"a}henb{\"u}hl, P. 2018.
\newblock Free supervision from video games.
\newblock In \emph{CVPR}, 2955--2964.

\bibitem[{Landrieu and Simonovsky(2018)}]{landrieu2018large}
Landrieu, L.; and Simonovsky, M. 2018.
\newblock Large-scale point cloud semantic segmentation with superpoint graphs.
\newblock In \emph{CVPR}, 4558--4567.

\bibitem[{Le and Duan(2018)}]{le2018pointgrid}
Le, T.; and Duan, Y. 2018.
\newblock Pointgrid: A deep network for 3d shape understanding.
\newblock In \emph{CVPR}, 9204--9214.

\bibitem[{Lee et~al.(2019)Lee, Kim, Kim, and Jeong}]{lee2019drop}
Lee, S.; Kim, D.; Kim, N.; and Jeong, S.-G. 2019.
\newblock Drop to adapt: Learning discriminative features for unsupervised
  domain adaptation.
\newblock In \emph{ICCV}, 91--100.

\bibitem[{Lei, Akhtar, and Mian(2019)}]{lei2019octree}
Lei, H.; Akhtar, N.; and Mian, A. 2019.
\newblock Octree guided CNN with spherical kernels for 3D point clouds.
\newblock In \emph{CVPR}, 9631--9640.

\bibitem[{Li, Chen, and Hee~Lee(2018)}]{li2018so}
Li, J.; Chen, B.~M.; and Hee~Lee, G. 2018.
\newblock So-net: Self-organizing network for point cloud analysis.
\newblock In \emph{CVPR}, 9397--9406.

\bibitem[{Li et~al.(2018{\natexlab{a}})Li, Bu, Sun, Wu, Di, and
  Chen}]{li2018pointcnn}
Li, Y.; Bu, R.; Sun, M.; Wu, W.; Di, X.; and Chen, B. 2018{\natexlab{a}}.
\newblock Pointcnn: Convolution on x-transformed points.
\newblock In \emph{NeurIPS}, 820--830.

\bibitem[{Li et~al.(2019)Li, Chen, Wang, and Zhang}]{li2019scale}
Li, Y.; Chen, Y.; Wang, N.; and Zhang, Z. 2019.
\newblock Scale-aware trident networks for object detection.
\newblock In \emph{ICCV}, 6054--6063.

\bibitem[{Li et~al.(2018{\natexlab{b}})Li, Wang, Shi, Hou, and
  Liu}]{li2018adaptivebn}
Li, Y.; Wang, N.; Shi, J.; Hou, X.; and Liu, J. 2018{\natexlab{b}}.
\newblock Adaptive Batch Normalization for practical domain adaptation.
\newblock \emph{PR} 80: 109--117.

\bibitem[{Lin et~al.(2020)Lin, Zhao, Meng, and Chua}]{lin2020multi}
Lin, C.; Zhao, S.; Meng, L.; and Chua, T.-S. 2020.
\newblock Multi-Source Domain Adaptation for Visual Sentiment Classification.
\newblock In \emph{AAAI}, 2661--2668.

\bibitem[{Lin et~al.(2017)Lin, Goyal, Girshick, He, and
  Doll{\'a}r}]{lin2017focal}
Lin, T.-Y.; Goyal, P.; Girshick, R.; He, K.; and Doll{\'a}r, P. 2017.
\newblock Focal loss for dense object detection.
\newblock In \emph{ICCV}, 2980--2988.

\bibitem[{Liu et~al.(2019{\natexlab{a}})Liu, Fan, Meng, Lu, Xiang, and
  Pan}]{liu2019densepoint}
Liu, Y.; Fan, B.; Meng, G.; Lu, J.; Xiang, S.; and Pan, C. 2019{\natexlab{a}}.
\newblock DensePoint: Learning densely contextual representation for efficient
  point cloud processing.
\newblock In \emph{ICCV}, 5239--5248.

\bibitem[{Liu et~al.(2019{\natexlab{b}})Liu, Tang, Lin, and Han}]{liu2019point}
Liu, Z.; Tang, H.; Lin, Y.; and Han, S. 2019{\natexlab{b}}.
\newblock Point-Voxel CNN for efficient 3D deep learning.
\newblock In \emph{NeurIPS}, 963--973.

\bibitem[{Long et~al.(2015)Long, Cao, Wang, and Jordan}]{long2015learning}
Long, M.; Cao, Y.; Wang, J.; and Jordan, M. 2015.
\newblock Learning transferable features with deep adaptation networks.
\newblock In \emph{ICML}, 97--105.

\bibitem[{Lyu, Huang, and Zhang(2020)}]{lyu2020learning}
Lyu, Y.; Huang, X.; and Zhang, Z. 2020.
\newblock Learning to Segment 3D Point Clouds in 2D Image Space.
\newblock In \emph{CVPR}, 12252--12261.

\bibitem[{Mao, Wang, and Li(2019)}]{mao2019interpolated}
Mao, J.; Wang, X.; and Li, H. 2019.
\newblock Interpolated convolutional networks for 3d point cloud understanding.
\newblock In \emph{ICCV}, 1578--1587.

\bibitem[{Meng et~al.(2019)Meng, Gao, Lai, and Manocha}]{meng2019vv}
Meng, H.-Y.; Gao, L.; Lai, Y.-K.; and Manocha, D. 2019.
\newblock VV-Net: Voxel vae net with group convolutions for point cloud
  segmentation.
\newblock In \emph{ICCV}, 8500--8508.

\bibitem[{Milioto et~al.(2019)Milioto, Vizzo, Behley, and
  Stachniss}]{milioto2019rangenet++}
Milioto, A.; Vizzo, I.; Behley, J.; and Stachniss, C. 2019.
\newblock Rangenet++: Fast and accurate lidar semantic segmentation.
\newblock In \emph{IROS}, 4213--4220.

\bibitem[{Morerio, Cavazza, and Murino(2018)}]{morerio2018minimal}
Morerio, P.; Cavazza, J.; and Murino, V. 2018.
\newblock Minimal-Entropy Correlation Alignment for Unsupervised Deep Domain
  Adaptation.
\newblock In \emph{ICLR}, 1--15.

\bibitem[{Patel et~al.(2015)Patel, Gopalan, Li, and
  Chellappa}]{patel2015visual}
Patel, V.~M.; Gopalan, R.; Li, R.; and Chellappa, R. 2015.
\newblock Visual domain adaptation: A survey of recent advances.
\newblock \emph{IEEE SPM} 32(3): 53--69.

\bibitem[{Peng et~al.(2019)Peng, Bai, Xia, Huang, Saenko, and
  Wang}]{peng2019moment}
Peng, X.; Bai, Q.; Xia, X.; Huang, Z.; Saenko, K.; and Wang, B. 2019.
\newblock Moment matching for multi-source domain adaptation.
\newblock In \emph{ICCV}, 1406--1415.

\bibitem[{Qi et~al.(2017{\natexlab{a}})Qi, Su, Mo, and Guibas}]{qi2017pointnet}
Qi, C.~R.; Su, H.; Mo, K.; and Guibas, L.~J. 2017{\natexlab{a}}.
\newblock Pointnet: Deep learning on point sets for 3d classification and
  segmentation.
\newblock In \emph{CVPR}, 652--660.

\bibitem[{Qi et~al.(2017{\natexlab{b}})Qi, Yi, Su, and
  Guibas}]{qi2017pointnet++}
Qi, C.~R.; Yi, L.; Su, H.; and Guibas, L.~J. 2017{\natexlab{b}}.
\newblock Pointnet++: Deep hierarchical feature learning on point sets in a
  metric space.
\newblock In \emph{NeurIPS}, 5099--5108.

\bibitem[{Qin et~al.(2019)Qin, You, Wang, Kuo, and Fu}]{qin2019pointdan}
Qin, C.; You, H.; Wang, L.; Kuo, C.-C.~J.; and Fu, Y. 2019.
\newblock PointDAN: A multi-scale 3D domain adaption network for point cloud
  representation.
\newblock In \emph{NeurIPS}, 7190--7201.

\bibitem[{Richter et~al.(2016)Richter, Vineet, Roth, and
  Koltun}]{richter2016playing}
Richter, S.~R.; Vineet, V.; Roth, S.; and Koltun, V. 2016.
\newblock Playing for data: Ground truth from computer games.
\newblock In \emph{ECCV}, 102--118.

\bibitem[{Rist, Enzweiler, and Gavrila(2019)}]{rist2019cross}
Rist, C.~B.; Enzweiler, M.; and Gavrila, D.~M. 2019.
\newblock Cross-Sensor Deep Domain Adaptation for LiDAR Detection and
  Segmentation.
\newblock In \emph{IV}, 1535--1542.

\bibitem[{Russo et~al.(2018)Russo, Carlucci, Tommasi, and
  Caputo}]{russo2018source}
Russo, P.; Carlucci, F.~M.; Tommasi, T.; and Caputo, B. 2018.
\newblock From source to target and back: symmetric bi-directional adaptive
  gan.
\newblock In \emph{CVPR}, 8099--8108.

\bibitem[{Saleh et~al.(2019)Saleh, Abobakr, Attia, Iskander, Nahavandi, Hossny,
  and Nahvandi}]{saleh2019domain}
Saleh, K.; Abobakr, A.; Attia, M.; Iskander, J.; Nahavandi, D.; Hossny, M.; and
  Nahvandi, S. 2019.
\newblock Domain Adaptation for Vehicle Detection from Bird's Eye View LiDAR
  Point Cloud Data.
\newblock In \emph{ICCVW}, 1--8.

\bibitem[{Sankaranarayanan et~al.(2018)Sankaranarayanan, Balaji, Castillo, and
  Chellappa}]{sankaranarayanan2018generate}
Sankaranarayanan, S.; Balaji, Y.; Castillo, C.~D.; and Chellappa, R. 2018.
\newblock Generate to adapt: Aligning domains using generative adversarial
  networks.
\newblock In \emph{CVPR}, 8503--8512.

\bibitem[{Shrivastava et~al.(2017)Shrivastava, Pfister, Tuzel, Susskind, Wang,
  and Webb}]{shrivastava2017learning}
Shrivastava, A.; Pfister, T.; Tuzel, O.; Susskind, J.; Wang, W.; and Webb, R.
  2017.
\newblock Learning from simulated and unsupervised images through adversarial
  training.
\newblock In \emph{CVPR}, 2107--2116.

\bibitem[{Sun, Feng, and Saenko(2016)}]{sun2016return}
Sun, B.; Feng, J.; and Saenko, K. 2016.
\newblock Return of frustratingly easy domain adaptation.
\newblock In \emph{AAAI}, 2058--2065.

\bibitem[{Sun and Saenko(2016)}]{dcoral}
Sun, B.; and Saenko, K. 2016.
\newblock Deep CORAL: Correlation Alignment for Deep Domain Adaptation.
\newblock In \emph{ECCVW}, 443--450.

\bibitem[{Sun et~al.(2019)Sun, Tzeng, Darrell, and Efros}]{sun2019unsupervised}
Sun, Y.; Tzeng, E.; Darrell, T.; and Efros, A.~A. 2019.
\newblock Unsupervised Domain Adaptation through Self-Supervision.
\newblock arXiv:1909.11825v2 [cs.LG].

\bibitem[{Te et~al.(2018)Te, Hu, Zheng, and Guo}]{te2018rgcnn}
Te, G.; Hu, W.; Zheng, A.; and Guo, Z. 2018.
\newblock Rgcnn: Regularized graph cnn for point cloud segmentation.
\newblock In \emph{ACM MM}, 746--754.

\bibitem[{Tripathi et~al.(2019)Tripathi, Chandra, Agrawal, Tyagi, Rehg, and
  Chari}]{tripathi2019learning}
Tripathi, S.; Chandra, S.; Agrawal, A.; Tyagi, A.; Rehg, J.~M.; and Chari, V.
  2019.
\newblock Learning to generate synthetic data via compositing.
\newblock In \emph{CVPR}, 461--470.

\bibitem[{Tzeng et~al.(2017)Tzeng, Hoffman, Saenko, and
  Darrell}]{tzeng2017adversarial}
Tzeng, E.; Hoffman, J.; Saenko, K.; and Darrell, T. 2017.
\newblock Adversarial discriminative domain adaptation.
\newblock In \emph{CVPR}, 2962--2971.

\bibitem[{Ulyanov, Vedaldi, and Lempitsky(2016)}]{ulyanov2016instance}
Ulyanov, D.; Vedaldi, A.; and Lempitsky, V. 2016.
\newblock Instance normalization: The missing ingredient for fast stylization.
\newblock arXiv:1607.08022v3 [cs.CV].

\bibitem[{Wang et~al.(2019{\natexlab{a}})Wang, Wu, Wu, and
  Keutzer}]{wang2019latte}
Wang, B.; Wu, V.; Wu, B.; and Keutzer, K. 2019{\natexlab{a}}.
\newblock LATTE: accelerating lidar point cloud annotation via sensor fusion,
  one-click annotation, and tracking.
\newblock In \emph{ITSC}, 265--272.

\bibitem[{Wang et~al.(2019{\natexlab{b}})Wang, Huang, Hou, Zhang, and
  Shan}]{wang2019graph}
Wang, L.; Huang, Y.; Hou, Y.; Zhang, S.; and Shan, J. 2019{\natexlab{b}}.
\newblock Graph attention convolution for point cloud semantic segmentation.
\newblock In \emph{CVPR}, 10296--10305.

\bibitem[{Wang et~al.(2017)Wang, Liu, Guo, Sun, and Tong}]{wang2017cnn}
Wang, P.-S.; Liu, Y.; Guo, Y.-X.; Sun, C.-Y.; and Tong, X. 2017.
\newblock O-cnn: Octree-based convolutional neural networks for 3d shape
  analysis.
\newblock \emph{ACM TOG} 36(4): 1--11.

\bibitem[{Wang et~al.(2019{\natexlab{c}})Wang, Sun, Liu, Sarma, Bronstein, and
  Solomon}]{wang2019dynamic}
Wang, Y.; Sun, Y.; Liu, Z.; Sarma, S.~E.; Bronstein, M.~M.; and Solomon, J.~M.
  2019{\natexlab{c}}.
\newblock Dynamic graph cnn for learning on point clouds.
\newblock \emph{ACM TOG} 38(5): 1--12.

\bibitem[{Wu et~al.(2018{\natexlab{a}})Wu, Wan, Yue, and
  Keutzer}]{wu2018squeezeseg}
Wu, B.; Wan, A.; Yue, X.; and Keutzer, K. 2018{\natexlab{a}}.
\newblock Squeezeseg: Convolutional neural nets with recurrent crf for
  real-time road-object segmentation from 3d lidar point cloud.
\newblock In \emph{ICRA}, 1887--1893.

\bibitem[{Wu et~al.(2019)Wu, Zhou, Zhao, Yue, and Keutzer}]{wu2019squeezesegv2}
Wu, B.; Zhou, X.; Zhao, S.; Yue, X.; and Keutzer, K. 2019.
\newblock Squeezesegv2: Improved model structure and unsupervised domain
  adaptation for road-object segmentation from a lidar point cloud.
\newblock In \emph{ICRA}, 4376--4382.

\bibitem[{Wu and He(2018)}]{wu2018group}
Wu, Y.; and He, K. 2018.
\newblock Group normalization.
\newblock In \emph{ECCV}, 3--19.

\bibitem[{Wu et~al.(2018{\natexlab{b}})Wu, Han, Lin, Gokhan~Uzunbas, Goldstein,
  Nam~Lim, and Davis}]{wu2018dcan}
Wu, Z.; Han, X.; Lin, Y.-L.; Gokhan~Uzunbas, M.; Goldstein, T.; Nam~Lim, S.;
  and Davis, L.~S. 2018{\natexlab{b}}.
\newblock Dcan: Dual channel-wise alignment networks for unsupervised scene
  adaptation.
\newblock In \emph{ECCV}, 518--534.

\bibitem[{Xu et~al.(2020)Xu, Wu, Wang, Zhan, Vajda, Keutzer, and
  Tomizuka}]{xu2020squeezesegv3}
Xu, C.; Wu, B.; Wang, Z.; Zhan, W.; Vajda, P.; Keutzer, K.; and Tomizuka, M.
  2020.
\newblock SqueezeSegV3: Spatially-Adaptive Convolution for Efficient
  Point-Cloud Segmentation.
\newblock In \emph{ECCV}, 1--19.

\bibitem[{Xu et~al.(2018)Xu, Fan, Xu, Zeng, and Qiao}]{xu2018spidercnn}
Xu, Y.; Fan, T.; Xu, M.; Zeng, L.; and Qiao, Y. 2018.
\newblock Spidercnn: Deep learning on point sets with parameterized
  convolutional filters.
\newblock In \emph{ECCV}, 87--102.

\bibitem[{Yu and Koltun(2016)}]{Yu2016}
Yu, F.; and Koltun, V. 2016.
\newblock Multi-scale context aggregation by dilated convolutions.
\newblock In \emph{ICLR}, 1--13.

\bibitem[{Yue et~al.(2018)Yue, Wu, Seshia, Keutzer, and
  Sangiovanni-Vincentelli}]{yue2018LiDAR}
Yue, X.; Wu, B.; Seshia, S.~A.; Keutzer, K.; and Sangiovanni-Vincentelli, A.~L.
  2018.
\newblock A lidar point cloud generator: from a virtual world to autonomous
  driving.
\newblock In \emph{ICMR}, 458--464.

\bibitem[{Zhang, Hua, and Yeung(2019)}]{zhang2019shellnet}
Zhang, Z.; Hua, B.-S.; and Yeung, S.-K. 2019.
\newblock Shellnet: Efficient point cloud convolutional neural networks using
  concentric shells statistics.
\newblock In \emph{ICCV}, 1607--1616.

\bibitem[{Zhao et~al.(2018{\natexlab{a}})Zhao, Zhang, Wu, Moura, Costeira, and
  Gordon}]{zhao2018adversarial}
Zhao, H.; Zhang, S.; Wu, G.; Moura, J.~M.; Costeira, J.~P.; and Gordon, G.~J.
  2018{\natexlab{a}}.
\newblock Adversarial multiple source domain adaptation.
\newblock In \emph{NeurIPS}, 8559--8570.

\bibitem[{Zhao et~al.(2019{\natexlab{a}})Zhao, Li, Yue, Gu, Xu, Hu, Chai, and
  Keutzer}]{zhao2019multi}
Zhao, S.; Li, B.; Yue, X.; Gu, Y.; Xu, P.; Hu, R.; Chai, H.; and Keutzer, K.
  2019{\natexlab{a}}.
\newblock Multi-source Domain Adaptation for Semantic Segmentation.
\newblock In \emph{NeurIPS}, 7285--7298.

\bibitem[{Zhao et~al.(2019{\natexlab{b}})Zhao, Lin, Xu, Zhao, Guo, Krishna,
  Ding, and Keutzer}]{zhao2019cycleemotiongan}
Zhao, S.; Lin, C.; Xu, P.; Zhao, S.; Guo, Y.; Krishna, R.; Ding, G.; and
  Keutzer, K. 2019{\natexlab{b}}.
\newblock CycleEmotionGAN: Emotional Semantic Consistency Preserved CycleGAN
  for Adapting Image Emotions.
\newblock In \emph{AAAI}, 2620--2627.

\bibitem[{Zhao et~al.(2020{\natexlab{a}})Zhao, Wang, Zhang, Gu, Li, Song, Xu,
  Hu, Chai, and Keutzer}]{zhao2020multi}
Zhao, S.; Wang, G.; Zhang, S.; Gu, Y.; Li, Y.; Song, Z.; Xu, P.; Hu, R.; Chai,
  H.; and Keutzer, K. 2020{\natexlab{a}}.
\newblock Multi-source Distilling Domain Adaptation.
\newblock In \emph{AAAI}, 12975--12983.

\bibitem[{Zhao et~al.(2021)Zhao, Xiao, Guo, Yue, Yang, Krishna, Xu, and
  Keutzer}]{zhao2021curriculum}
Zhao, S.; Xiao, Y.; Guo, J.; Yue, X.; Yang, J.; Krishna, R.; Xu, P.; and
  Keutzer, K. 2021.
\newblock Curriculum CycleGAN for Textual Sentiment Domain Adaptation with
  Multiple Sources.
\newblock In \emph{WWW}, to appear.

\bibitem[{Zhao et~al.(2020{\natexlab{b}})Zhao, Yue, Zhang, Li, Zhao, Wu,
  Krishna, Gonzalez, Sangiovanni-Vincentelli, Seshia et~al.}]{zhao2020review}
Zhao, S.; Yue, X.; Zhang, S.; Li, B.; Zhao, H.; Wu, B.; Krishna, R.; Gonzalez,
  J.~E.; Sangiovanni-Vincentelli, A.~L.; Seshia, S.~A.; et~al.
  2020{\natexlab{b}}.
\newblock A Review of Single-Source Deep Unsupervised Visual Domain Adaptation.
\newblock \emph{IEEE TNNLS} to appear.

\bibitem[{Zhao et~al.(2018{\natexlab{b}})Zhao, Zhao, Ding, and
  Keutzer}]{zhao2018emotiongan}
Zhao, S.; Zhao, X.; Ding, G.; and Keutzer, K. 2018{\natexlab{b}}.
\newblock EmotionGAN: unsupervised domain adaptation for learning discrete
  probability distributions of image emotions.
\newblock In \emph{ACM MM}, 1319--1327.

\bibitem[{Zhu et~al.(2017)Zhu, Park, Isola, and Efros}]{zhu2017unpaired}
Zhu, J.-Y.; Park, T.; Isola, P.; and Efros, A.~A. 2017.
\newblock Unpaired Image-To-Image Translation Using Cycle-Consistent
  Adversarial Networks.
\newblock In \emph{ICCV}, 2223--2232.

\bibitem[{Zhuo et~al.(2017)Zhuo, Wang, Zhang, and Huang}]{zhuo2017deep}
Zhuo, J.; Wang, S.; Zhang, W.; and Huang, Q. 2017.
\newblock Deep Unsupervised Convolutional Domain Adaptation.
\newblock In \emph{ACM MM}, 261--269.

\end{thebibliography}

\end{document}